\documentclass{article}
\usepackage[final]{neurips_2022}

\usepackage{hyperref}
\usepackage{color,xcolor}
\usepackage{epsfig}
\usepackage{graphicx}
\usepackage{subcaption}

\usepackage{tikz}
\usetikzlibrary{arrows}

\usepackage{listings}
\usepackage{esdiff}
\usepackage{mdframed}
\usepackage{lipsum}
\newmdenv[
  backgroundcolor=gray!20,
  skipabove=\topsep,
  skipbelow=\topsep,
  roundcorner=2mm
]{gentext}

\usepackage{pifont}
\newcommand{\cmark}{\ding{51}}%
\newcommand{\xmark}{\ding{55}}%

\usepackage{tabularx}
\usepackage{adjustbox}
\usepackage{array}
\usepackage{booktabs}
\usepackage{colortbl}
\usepackage{float,wrapfig}
\usepackage{hhline}
\usepackage{multirow}
\usepackage{numprint}

\usepackage{amsmath,amsfonts,amsthm,amssymb}
\usepackage{mathtools}
\usepackage{bm}
\usepackage{nicefrac}
\usepackage{microtype}
\usepackage{dsfont}
\usepackage{changepage}
\usepackage{extramarks}
\usepackage{fancyhdr}
\usepackage{lastpage}
\usepackage{setspace}
\usepackage{soul}
\usepackage{xspace}

\usepackage{algorithm, algpseudocode}
\usepackage{enumitem}
\usepackage[textsize=tiny]{todonotes} %

\usepackage{titlesec}

\usepackage[page]{appendix}
\usepackage[hang,flushmargin]{footmisc}

\titlespacing*\section{0pt}{12pt plus 5pt minus 5pt}{9pt plus 3pt minus 3pt}

\hypersetup{
    colorlinks,
    linkcolor={red!50!black},
    citecolor={blue!50!black},
    urlcolor={blue!80!black}
}

\usepackage{amsmath,amsfonts,bm}
\usepackage{xifthen}
\usepackage{xparse}

\catcode`\_=11\relax

\def\_email#1@#2\q_nil{%
  \href{mailto:#1@#2}{{\emailfont #1\emailampersat #2}}
}
\newcommand\emailfont{\rmfamily \lsstyle}
\newcommand\emailampersat{@}
\catcode`\_=8\relax

\definecolor{ggreen}{rgb}{0.0, 0.6, 0.0}
\definecolor{rred}{rgb}{0.75, 0.0, 0.0}
\definecolor{bblue}{rgb}{0.13, 0.67, 0.8}

\newcommand{\badmetric}[1]{{\color{rred} \textbf{#1}}}
\newcommand{\goodmetric}[1]{{\color{ggreen} \textbf{#1}}}

\newcommand{\rewg}{G^\prime}  %
\newcommand{\mlp}{\mathrm{mlp}\xspace}

\newcommand{\cfd}{\textsc{CounterFact}\xspace}
\newcommand{\cfdabbr}{\textsc{CF}\xspace}
\newcommand{\pararel}{\textsc{ParaRel}\xspace}
\newcommand{\efk}{Knowledge Editor\xspace}
\newcommand{\dkn}{Knowledge Neurons\xspace}
\newcommand{\methodlong}{Rank-One Model Editing\xspace}
\newcommand{\method}{ROME\xspace}

\newcommand{\atl}[2]{#1^{(#2)}}

\newcommand{\pr}[1]{\mathbb{P}\left[ #1 \right]}
\newcommand{\prsub}[2]{\mathbb{P}_{#2}\left[ #1 \right]}

\newcommand{\reffig}[1]{Figure~\ref{fig:#1}}
\newcommand{\refsec}[1]{Section~\ref{sec:#1}}

\newcommand{\refeqn}[1]{Eqn.~\ref{eq:#1}}

\newcommand{\lblfig}[1]{\label{fig:#1}}

\newcommand{\lbleq}[1]{\label{eq:#1}}

\newcommand{\ignorethis}[1]{}

\def\eqref#1{equation~\ref{#1}}

\def\1{\bm{1}}

\DeclareMathAlphabet{\mathsfit}{\encodingdefault}{\sfdefault}{m}{sl}
\SetMathAlphabet{\mathsfit}{bold}{\encodingdefault}{\sfdefault}{bx}{n}

\DeclareMathOperator*{\argmin}{argmin}

\newcolumntype{L}[1]{>{\raggedright\let\newline\\\arraybackslash\hspace{0pt}}m{#1}}
\newcolumntype{C}[1]{>{\centering\let\newline\\\arraybackslash\hspace{0pt}}m{#1}}
\newcolumntype{R}[1]{>{\raggedleft\let\newline\\\arraybackslash\hspace{0pt}}m{#1}}

\newcommand{\ignore}[1]{}

\makeatletter
\DeclareRobustCommand\onedot{\futurelet\@let@token\@onedot}
\def\@onedot{\ifx\@let@token.\else.\null\fi\xspace}

\makeatother

\definecolor{MyDarkBlue}{rgb}{0,0.08,1}
\definecolor{MyDarkGreen}{rgb}{0.02,0.6,0.02}
\definecolor{MyDarkRed}{rgb}{0.8,0.02,0.02}
\definecolor{MyDarkOrange}{rgb}{0.40,0.2,0.02}
\definecolor{MyPurple}{RGB}{111,0,255}
\definecolor{MyRed}{rgb}{1.0,0.0,0.0}
\definecolor{MyGold}{rgb}{0.75,0.6,0.12}
\definecolor{MyDarkgray}{rgb}{0.66, 0.66, 0.66}

\usepackage{amsmath,amsfonts,bm}

\DeclareMathAlphabet{\mathsfit}{\encodingdefault}{\sfdefault}{m}{sl}
\SetMathAlphabet{\mathsfit}{bold}{\encodingdefault}{\sfdefault}{bx}{n}

\title{Locating and Editing Factual Associations in GPT}

\author{%
Kevin Meng\thanks{Equal contribution. Correspondence to \texttt{mengk@mit.edu}, \texttt{davidbau@northeastern.edu}.} \\
MIT CSAIL \\
\And
David Bau$^*$ \\
Northeastern University \\
\And
Alex Andonian \\
MIT CSAIL \\
\And
Yonatan Belinkov\thanks{Supported by the Viterbi Fellowship in the Center for Computer Engineering at the Technion.} \\
Technion -- IIT \\
}

\begin{document}

\maketitle

\addtolength{\tabcolsep}{-5pt}

\begin{abstract}
We analyze the storage and recall of factual associations in autoregressive transformer language models, finding evidence that these associations correspond to localized, directly-editable computations. We first develop a causal intervention for identifying neuron \textit{activations} that are decisive in a model's factual predictions. This reveals a distinct set of steps in middle-layer feed-forward modules that mediate factual predictions while processing subject tokens. To test our hypothesis that these computations correspond to factual association recall, we modify feed-forward \textit{weights} to update specific factual associations using Rank-One Model Editing (ROME).  We find that ROME is effective on a standard zero-shot relation extraction (zsRE) model-editing task. We also evaluate ROME on a new dataset of difficult counterfactual assertions, on which it simultaneously maintains both specificity and generalization, whereas other methods sacrifice one or another. Our results confirm an important role for mid-layer feed-forward modules in storing factual associations and suggest that direct manipulation of computational mechanisms may be a feasible approach for model editing. The code, dataset, visualizations, and an interactive demo notebook are available
at \url{https://rome.baulab.info/}.

\end{abstract}
\section{Introduction}

Where does a large language model store its facts? In this paper, we report evidence that factual associations in GPT correspond to a localized computation that can be directly edited.

Large language models can predict factual statements about the world~\citep{petroni-etal-2019-language,jiang-etal-2020-know,roberts-etal-2020-much}. For example, given the prefix ``\emph{The Space Needle is located in the city of},'' GPT will reliably predict the true answer: ``\emph{Seattle}'' (\reffig{infoflow}a).  Factual knowledge has been observed to emerge in both autoregressive GPT models~\citep{gpt2,gpt3} and masked BERT models~\citep{devlin-etal-2019-bert}.

In this paper, we investigate how such factual associations are stored within GPT-like autoregressive transformer models. Although many of the largest neural networks in use today are autoregressive, the way that they store knowledge remains under-explored. Some research has been done for masked models~\citep{petroni-etal-2019-language,jiang-etal-2020-know,10.1162/tacl_a_00410,geva-etal-2021-transformer,dai-kn,decao-ke}, but GPT has architectural differences such as unidirectional attention and generation capabilities that provide an opportunity for new insights.

We use two approaches. First, we trace the causal effects of hidden state \underline{activations} within GPT using causal mediation analysis~\citep{pearl2001direct,vig2020investigating} to identify the specific modules that mediate recall of a fact about a subject (\reffig{infoflow}).  Our analysis reveals that feedforward MLPs at a range of middle layers are decisive when processing the last token of the subject name (Figures~\ref{fig:infoflow}b,\ref{fig:avg-1000-trace}b,\ref{fig:trace-mlp-disabled}).

Second, we test this finding in model \underline{weights} by introducing a Rank-One Model Editing method (ROME) to alter the parameters that determine a feedfoward layer's behavior at the decisive token.  Despite the simplicity of the intervention, we find that ROME is similarly effective to other model-editing approaches on a standard zero-shot relation extraction benchmark (Section~\ref{subsec:zsre}).

To evaluate ROME's impact on more difficult cases, we introduce a dataset of counterfactual assertions (Section~\ref{subsec:counterfact-description}) that would not have been observed in pretraining. Our evaluations (Section~\ref{subsec:romeeval-mlp}) confirm that midlayer MLP modules can store factual associations that generalize beyond specific surface forms, while remaining specific to the subject. Compared to previous fine-tuning~\citep{zhu-ft}, interpretability-based~\citep{dai-kn}, and meta-learning~\citep{mend,decao-ke} methods, ROME achieves good generalization and specificity simultaneously, whereas previous approaches sacrifice one or the other.

\begin{figure*}%
\includegraphics[width=\textwidth]{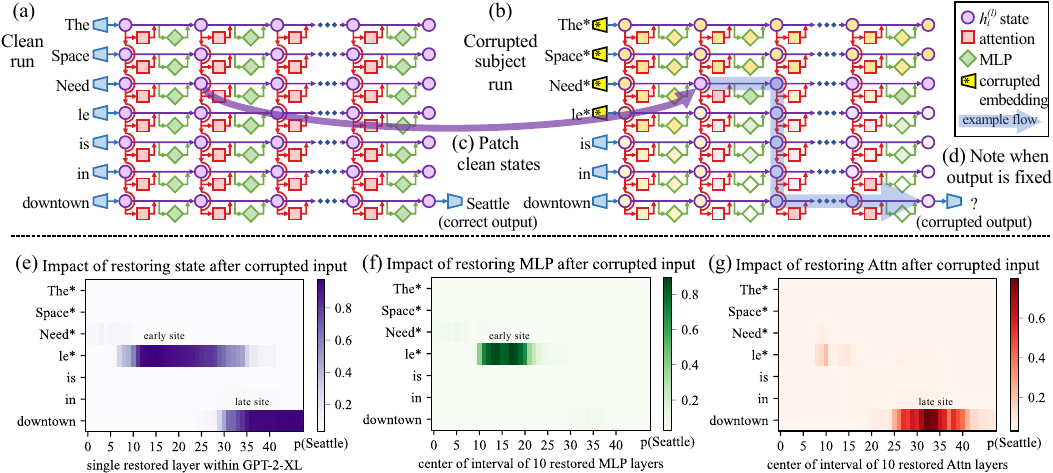}%
\caption{\small \textbf{Causal Traces} compute the causal effect of neuron activations by running the network twice: (a) once normally, and (b) once where we corrupt the subject token and then (c) restore selected internal activations to their clean value. (d) Some sets of activations cause the output to return to the original prediction; the light blue path shows an example of information flow.  The causal impact on output probability is mapped for the effect of (e) each hidden state on the prediction, (f) only MLP activations, and (g) only attention activations.}%
\lblfig{infoflow}%
\end{figure*}%

\section{Interventions on \texorpdfstring{\underline{Activations}}{Activations} for Tracing Information Flow}
\label{sec:causal-tracing}

To locate facts within the parameters of a large pretrained autoregressive transformer, we begin by analyzing and identifying the specific hidden states that have the strongest causal effect on predictions of individual facts.
We represent each fact as a knowledge tuple $t = \left( s, r, o \right)$ containing the subject $s$, object $o$, and relation $r$ connecting the two.
Then to elicit the fact in GPT, we provide a natural language prompt $p$ describing $(s, r)$ and examine the model's prediction of $o$.

An autoregressive transformer language model $G: \mathcal{X} \rightarrow \mathcal{Y}$ over vocabulary $V$ maps a token sequence $x = [x_1, ..., x_T] \in \mathcal{X}$, $x_i \in V$ to a probability distribution $\smash{y \in \mathcal{Y} \subset \mathbb{R}^{|V|}}$ that predicts next-token continuations of $x$.  Within the transformer, the $i$th token is embedded as a series of hidden state vectors $\smash{\atl{h}{l}_i}$, beginning with $\smash{\atl{h}{0}_i = \mathrm{emb}(x_i) + \mathrm{pos}(i) \in \mathbb{R}^{H}}$. The final output $\smash{y = \mathrm{decode}(\atl{h}{L}_{T})}$ is read from the last hidden state.

We visualize the internal computation of $G$ as a grid (\reffig{infoflow}a) of hidden states $\smash{\atl{h}{l}_i}$ in which each layer $l$ (left $\rightarrow$ right) adds global attention $\smash{\atl{a}{l}_i}$ and local MLP $\smash{\atl{m}{l}_i}$ contributions computed from previous layers, and where each token $i$ (top $\rightarrow$ bottom) attends to previous states from other tokens.  Recall that, in the autoregressive case, tokens only draw information from past (above) tokens:
\begin{align}
    \notag
    \atl{h}{l}_i = \atl{h}{l-1}_i + \atl{a}{l}_i &+ \atl{m}{l}_i
     \\
    \label{eq:autoregressive}
    \atl{a}{l}_i &= \atl{\mathrm{attn}}{l}\left(\atl{h}{l-1}_1, \atl{h}{l-1}_2, \dots, \atl{h}{l-1}_i\right) \\
    \notag
    \atl{m}{l}_i &=  \atl{W_{proj}}{l}\,
    \sigma\left(  \atl{W_{fc}}{l} \gamma\left( \atl{a}{l}_i + \atl{h}{l-1}_i \right) \right).
\end{align}
Each layer's MLP is a two-layer neural network parameterized by matrices $\smash{\atl{W_{proj}}{l}}$ and $\smash{\atl{W_{fc}}{l}}$, with rectifying nonlinearity $\sigma$ and normalizing nonlinearity $\gamma$. For further background on transformers, we refer to \cite{vaswani2017attention}.\footnote{\refeqn{autoregressive} calculates attention sequentially after the MLP module as in \citet{gpt3}. Our methods also apply to GPT variants such as \citet{gpt-j} that put attention in parallel to the MLP.}

\begin{figure*}%
\centering%
\includegraphics[width=\textwidth]{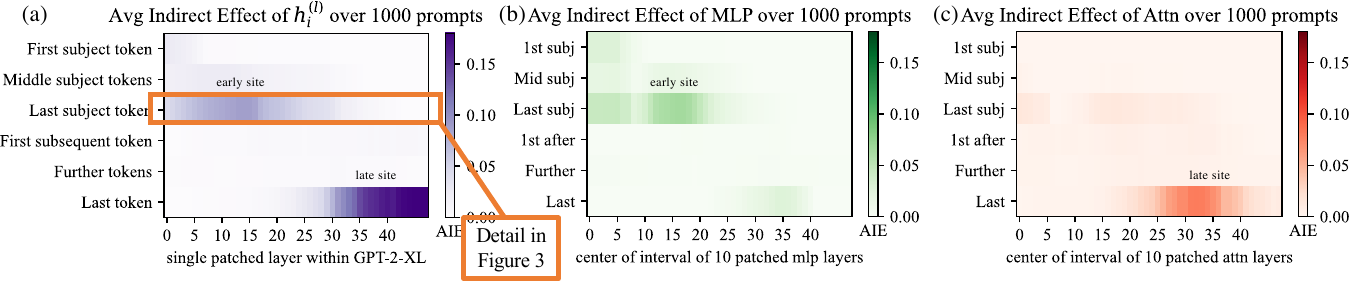}%
\caption{\small \textbf{Average Indirect Effect} of individual model components over a sample of 1000 factual statements reveals two important sites.  (a) Strong causality at a `late site' in the last layers at the last token is unsurprising, but strongly causal states at an `early site' in middle layers at the last subject token is a new discovery.  (b) MLP contributions dominate the early site. (c) Attention is important at the late site. Appendix~\ref{apd:causal-tracing}, \reffig{apd-ct-line-plots} shows these heatmaps as line plots with 95\% confidence intervals.}%
\lblfig{avg-1000-trace}%
\end{figure*}%

\subsection{Causal Tracing of Factual Associations} \label{subsec:cma}

The grid of states (\reffig{infoflow})
forms a %
\emph{causal graph}~\citep{pearl2009} describing dependencies between the hidden variables.  This graph contains many paths from inputs on the left to the output (next-word prediction) at the lower-right, and we wish to understand if there are specific hidden state variables that are more important than others when recalling a fact.

As \citet{vig2020investigating} have shown, this is a natural case for \textit{causal mediation analysis}, which quantifies the contribution of intermediate variables in causal graphs \citep{pearl2001direct}. To calculate each state's contribution towards a correct factual prediction, we observe all of $G$'s internal activations during three runs: a \textbf{clean} run that predicts the fact, a \textbf{corrupted} run where the prediction is damaged, and a \textbf{corrupted-with-restoration} run that tests the ability of a single state to restore the prediction.

\begin{itemize}[leftmargin=*,topsep=0pt,parsep=0pt,partopsep=0pt,itemsep=3pt]
    \item In the \textbf{clean run}, we pass a factual prompt $x$ into $G$ and collect all hidden activations $\left\{ \smash{\atl{h}{l}_i} \mid i \in[1,T], l\in[1,L] \right\}$. \reffig{infoflow}a provides an example illustration with the prompt: ``The Space Needle is in downtown \rule{1cm}{0.15mm}'', for which the expected completion is $o = $ ``Seattle''.
    
    \item In the baseline \textbf{corrupted run}, the subject is obfuscated from $G$ before the network runs. Concretely, immediately after $x$ is embedded as $\left[ \smash{\atl{h}{0}_{1}}, \smash{\atl{h}{0}_{2}}, \dots, \smash{\atl{h}{0}_{T}} \right]$, we set $\smash{\atl{h}{0}_{i} := \atl{h}{0}_{i}} + \epsilon$ for all indices $i$ that correspond to the subject entity, where $\epsilon \sim \mathcal{N}(0;\nu)$\footnote{We select $\nu$ to be $3$ times larger than the empirical standard deviation of embeddings; see Appendix~\ref{apd:causal-tracing-details} for details, and see Appendix~\ref{apd:causal-tracing-variations} for an analysis of other corruption rules.}; . $G$ is then allowed to continue normally, giving us a set of corrupted activations $\left\{ \smash{\atl{h}{l}_{i*}} \mid i \in[1,T], l\in[1,L] \right\}$. Because $G$ loses some information about the subject, it will likely return an incorrect answer (\reffig{infoflow}b).
    
    \item The \textbf{corrupted-with-restoration run}, lets $G$ run computations on the noisy embeddings as in the corrupted baseline, \textit{except} at some token $\smash{\hat{i}}$ and layer $\smash{\hat{l}}$. There, we hook $G$ so that it is forced to output the clean state $\smash{\atl{h}{\smash{\hat{l}}}_{\hat{i}}}$; future computations execute without further intervention. Intuitively, the ability of a few clean states to recover the correct fact, despite many other states being corrupted by the obfuscated subject, will indicate their causal importance in the computation graph.
\end{itemize} 

Let $\mathbb{P}[o]$, $\mathbb{P}_*[o]$, and $\mathbb{P}_{*,\smash{\text{ clean } \atl{h_i}{l}}}[o]$ denote the probability of emitting $o$ under the clean, corrupted, and corrupted-with-restoration runs, respectively; dependence on the input $x$ is omitted for notational simplicity.
The \textbf{total effect} (TE) is the difference between these quantities:  $ \text{TE} = \mathbb{P}[o] - \mathbb{P}_*[o]$. The \textbf{indirect effect} (IE) of a specific mediating state $\smash{\atl{h_i}{l}}$ is defined as the difference between the probability of $o$ under the corrupted version and the probability when that state is set to its clean version, while the subject remains corrupted: $\text{IE} =  \mathbb{P}_{*,\smash{\text{ clean } \atl{h_i}{l}}}[o] - \mathbb{P}_*[o]$. 
Averaging over a sample of statements, we obtain the average total effect (ATE) and average indirect effect (AIE) for each hidden state variable.\footnote{One could also compute the direct effect, which flows through other model components besides the chosen mediator. However, we found this effect to be noisy and uninformative, in line with results by \citet{vig2020investigating}.}

\subsection{Causal Tracing Results} 
We compute the average indirect effect (AIE) over 1000 factual statements (details in Appendix~\ref{apd:causal-tracing-details}), varying the mediator over different positions in the sentence and different model components including individual states, MLP layers, and attention layers. 
\reffig{avg-1000-trace} plots the AIE of the internal components of GPT-2~XL (1.5B parameters).  The ATE of this experiment is 18.6\%, and we note that a large portion of the effect is mediated by strongly causal individual states (AIE=8.7\% at layer 15) at the last subject token.  The presence of strong causal states at a late site immediately before the prediction is unsurprising, but their emergence at an \textit{early} site at the last token of the subject is a new discovery.

Decomposing the causal effects of contributions of MLP and attention modules (Figure~\ref{fig:infoflow}fg and Figure~\ref{fig:avg-1000-trace}bc) suggests a decisive role for MLP modules at the early site: MLP contributions peak at AIE 6.6\%, while attention at the last subject token is only AIE 1.6\%; attention is more important at the last token of the prompt. Appendix~\ref{apd:mlp-and-attn-trace} further discusses this decomposition.

Finally, to gain a clearer picture of the special role of MLP layers at the early site, we analyze indirect effects with a modified causal graph (\reffig{trace-mlp-disabled}).  (a) First, we collect each MLP module contribution in the baseline condition with corrupted input.  (b) Then, to isolate the effects of MLP modules when measuring causal effects, we modify the computation graph to sever MLP computations at token $i$ and freeze them in the baseline corrupted state so that they are unaffected by the insertion of clean state for $\smash{\atl{h_i}{l}}$.  This modification is a way of probing \emph{path-specific effects}~\citep{pearl2001direct} for paths that avoid MLP computations.  (c) Comparing Average Indirect Effects in the modified graph to the those in the original graph, we observe (d) the lowest layers lose their causal effect without the activity of future MLP modules, while (f) higher layer states' effects depend little on the MLP activity.  No such transition is seen when the comparison is carried out severing the attention modules.  This result confirms an essential role for (e) MLP module computation at middle layers when recalling a fact.

Appendix~\ref{apd:causal-tracing} has results on other autoregressive models and experimental settings. In particular, we find that Causal Tracing is more informative than gradient-based salience methods such as integrated gradients \citep{sundararajan2017axiomatic} (\reffig{apd-ig}) and is robust under different noise configurations.

\begin{figure}%
\centering %
\includegraphics[width=\textwidth]{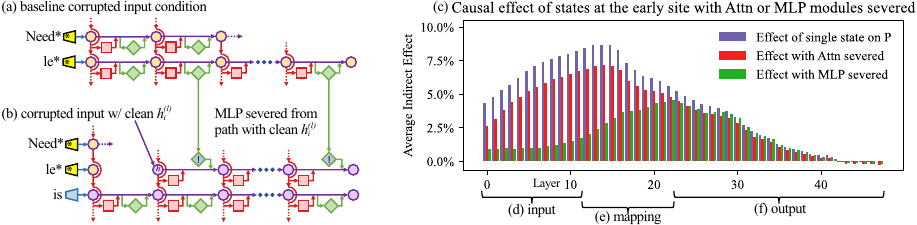}%
\caption{\small \textbf{Causal effects with a modified computation graph}.  (a,b) To isolate the effects of MLP modules when measuring causal effects, the computation graph is modified.  (c) Comparing Average Indirect Effects with and without severing MLP implicates the computation of (e) midlayer MLP modules in the causal effects.  No similar gap is seen when attention is similarly severed.}%
\lblfig{trace-mlp-disabled}%
\vspace{-5pt}%
\end{figure}

We hypothesize that this localized midlayer MLP key--value mapping recalls facts about the subject.

\subsection{The Localized Factual Association Hypothesis}

Based on causal traces, we posit a specific mechanism for storage of factual associations: each midlayer MLP module accepts inputs that encode a subject, then produces outputs that recall memorized properties about that subject. Middle layer MLP outputs accumulate information, then the summed information is copied to the last token by attention at high layers.

This hypothesis localizes factual association along three dimensions, placing it (i) in the MLP modules (ii) at specific middle layers (iii) and specifically at the processing of the subject's last token.  It is consistent with the~\citet{geva-etal-2021-transformer} view that MLP layers store knowledge, and the~\citet{anthropic2021} study showing an information-copying role for self-attention.  Furthermore, informed  by the \citet{zhao2021non} finding that transformer layer order can be exchanged with minimal change in behavior, we propose that this picture is complete. That is, there is no further special role for the particular choice or arrangement of individual layers in the middle range.  We conjecture that any fact could be equivalently stored in any one of the middle MLP layers.
To test our hypothesis, we narrow our attention to a single MLP module at a mid-range layer $l^*$, and ask whether its weights can be explicitly modified to store an arbitrary fact.

\section{Interventions on \texorpdfstring{\underline{Weights}}{Weights} for Understanding Factual Association Storage} \label{sec:rome}

While Causal Tracing has implicated MLP modules in recalling factual associations, we also wish to understand how facts are \textit{stored in weights}. \citet{geva-etal-2021-transformer} observed that MLP layers (\reffig{rome-method}cde) can act as two-layer key--value memories,\footnote{Unrelated to keys and values in self-attention.} where the neurons of the first layer $\smash{\atl{W}{l}_{fc}}$ form a \textit{key}, with which the second layer $\smash{\atl{W}{l}_{proj}}$ retrieves an associated \textit{value}. We hypothesize that MLPs can be modeled as a linear associative memory; note that this differs from \citeauthor{geva-etal-2021-transformer}'s per-neuron view.

We test this hypothesis by conducting a new type of intervention: modifying factual associations with \methodlong (\method). Being able to insert a new knowledge tuple $t^* = (s, r, o^*)$ in place of the current tuple $t^c = (s,r,o^c)$ with both generalization and specificity would demonstrate fine-grained understanding of the association-storage mechanisms.

\subsection{\methodlong: Viewing the Transformer MLP as an Associative Memory}
\label{subsec:rome-math}
\begin{figure}[t]
\includegraphics[width=1\columnwidth]{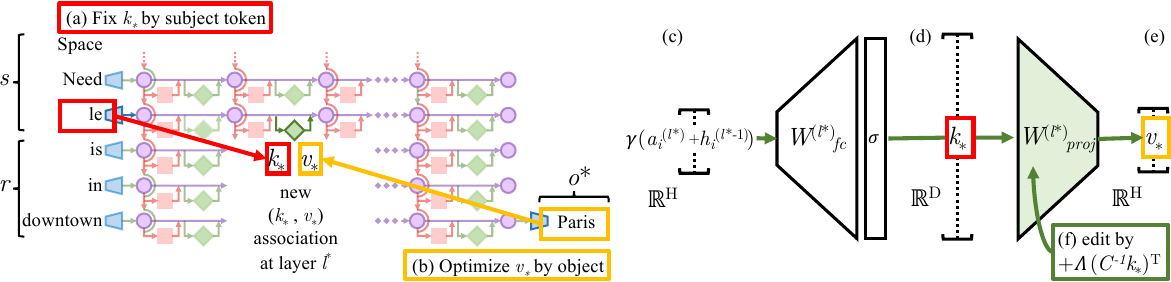}%
\caption{\small \textbf{Editing one MLP layer with \method}.  To associate \emph{Space Needle} with \emph{Paris}, the \method method inserts a new $(k_*, v_*)$ association into layer $l^*$, where (a) key $k_*$ is determined by the subject and (b) value $v_*$ is optimized to select the object.  (c) Hidden state at layer $l^*$ and token $i$ is expanded to produce (d) the key vector $k_*$ for the subject. (e) To write new value vector $v_*$ into the layer, (f) we calculate a rank-one update $\Lambda(C^{-1}k_*)^T$ to cause $\smash{\atl{\hat{W}}{l}_{proj} k_* = v_*}$ while minimizing interference with other memories stored in the layer.}%
\lblfig{rome-method}%
\end{figure}

We view $\smash{\atl{W}{l}_{proj}}$ as a linear associative memory~\citep{kohonen1972correlation,anderson1972simple}.  This perspective observes that any linear operation $W$ can operate as a key--value store for a set of vector keys $K = [ k_1 \mid k_2 \mid \dots ]$ and corresponding vector values $V = [v_1 \mid v_2 \mid \dots]$, by solving $WK \approx V$, whose squared error is minimized using the Moore-Penrose pseudoinverse: $W = VK^{+}$.
\citet{bau-rewriting} observed that a new key--value pair $(k_*, v_*)$ can be inserted optimally into the memory by solving a constrained least-squares problem. In a convolutional network, \citeauthor{bau-rewriting} solve this using an optimization, but in a fully-connected layer, we can derive a closed form solution:
\begin{align}
\text{minimize} \; \lVert \hat{W}K &- V \rVert \; \text{such that} \;  \hat{W}k_* = v_* \quad \text{by setting} \; \hat{W} = W + \Lambda (C^{-1}k_*)^T.
\lbleq{cinvk}
\end{align}
Here $W$ is the original matrix, $C=KK^T$ is a constant that we pre-cache by estimating the uncentered covariance of $k$ from a sample of Wikipedia text (Appendix~\ref{subapd:rome-hparams}), and $\Lambda = (v_* - W k_*)/(C^{-1}k_*)^T k_*$ is a vector proportional to the residual error of the new key--value pair on the original memory matrix (full derivation in Appendix \ref{apd:solving-v}).
Because of this simple algebraic structure, we can insert any fact directly once $(k_*, v_*)$ is computed. All that remains is to choose the appropriate $k_*$ and $v_*$.

\textbf{Step 1: Choosing \texorpdfstring{$k_*$}{k*} to Select the Subject.}
Based on the decisive role of MLP inputs at the final subject token (Section \ref{sec:causal-tracing}), we shall choose inputs that represent the subject at its last token as the lookup key $k_*$.
Specifically, we compute $k_*$ by collecting activations: We pass text $x$ containing the subject $s$ through $G$; then at layer $l^*$ and last subject token index $i$, we read the value after the non-linearity inside the MLP (\reffig{rome-method}d). Because the state will vary depending on tokens that precede $s$ in text, we set $k_*$ to an average value over a small set of texts ending with the subject $s$:
\begin{align}
k_* = \frac{1}{N} \sum_{j=1}^N k(x_j + s), \; \text{where} \; k(x) = \sigma\left( \atl{W_{fc}}{l^*} \; \gamma(\atl{a}{l^*}_{[x],i} + \atl{h}{l^*-1}_{[x],i})  \right).
\lbleq{kstar-sample}
\end{align}
In practice, we sample $x_j$ by generating 50 random token sequences of length 2 to 10 using $G$.

\textbf{Step 2: Choosing \texorpdfstring{$v_*$}{v*} to Recall the Fact.}
Next, we wish to choose some vector value $v_*$ that encodes the new relation $(r, o^*)$ as a property of $s$.
We set $v_* = \argmin_z \mathcal{L}(z)$, where the objective $\mathcal{L}(z)$ is:
\begin{align}\label{eq:v-optimization}
\frac{1}{N} \sum_{j=1}^N \underbrace{-\log\prsub{o^* \mid x_j + p\,}{G(\atl{m}{l^*}_{i}:=z)}}_\text{(a) Maximizing $o^*$ probability} \; + \; \underbrace{D_{\mathrm{KL}}\left(\prsub{x \mid p^\prime}{G(\atl{m}{l^*}_{i^\prime}:=z)} \big\Vert \prsub{x \mid p^\prime}{G} \right)}_\text{(b) Controlling essence drift}.
\end{align}
The first term (Eqn. \ref{eq:v-optimization}a) seeks a vector $z$ that, when substituted as the output of the MLP at the token $i$ at the end of the subject (notated $\smash{G(\atl{m}{l^*}_{i}:=z)}$), will cause the network to predict the target object $o^*$ in response to the factual prompt $p$.
The second term (Eqn. \ref{eq:v-optimization}b) minimizes the KL divergence of predictions for the prompt $p^\prime$ (of the form ``\{subject\} is a'') to the unchanged model, which helps preserve the model's understanding of the subject's essence.
To be clear, the optimization does \textit{not} directly alter model weights; it identifies a vector representation $v_*$ that, when output at the targeted MLP module, represents the new property $(r, o^*)$ for the subject $s$.
Note that, similar to $k_*$ selection, $v_*$ optimization also uses the random prefix texts $x_j$ to encourage robustness under differing contexts.

\textbf{Step 3: Inserting the Fact.}
Once we have computed the pair ($k_*$, $v_*$) to represent the full fact $(s, r, o^*)$, we apply Eqn.~\ref{eq:cinvk}, updating the MLP weights $\smash{\atl{W}{l}_{proj}}$ with a rank-one update that inserts the new key--value association directly.
For full implementation details, see Appendix \ref{subapd:rome-hparams}.

\subsection{Evaluating \method: Zero-Shot Relation Extraction (zsRE)} \label{subsec:zsre}

We wish to test our localized factual association hypothesis: can storing a single new vector association using \method insert a substantial, generalized factual association into the model?

A natural question is how \method compares to other model-editing methods, which use direct optimization or hypernetworks to incorporate a single new training example into a network. For baselines, we examine Fine-Tuning \textbf{(FT)}, which applies Adam with early stopping at one layer to minimize $-\log \pr{o^* \mid x}$.
Constrained Fine-Tuning \textbf{(FT+L)} \citep{zhu-ft} additionally imposes a parameter-space $L_\infty$ norm constraint on weight changes.
We also test two hypernetworks: \efk \textbf{(KE)} \citep{decao-ke} and \textbf{MEND} \citep{mend}, both of which learn auxiliary models to predict weight changes to $G$. Further details are described in Appendix~\ref{apd:implementations}. 

\begin{wraptable}{R}{0.5\textwidth}
\vspace{-13pt}%
    \centering
    \caption{\small zsRE Editing Results on GPT-2 XL.}
    \vspace{-5pt}%
    \label{tab:zsre-results}    
    \tiny
    \begin{adjustbox}{width=0.5\textwidth}
    \begin{tabular}{l r r r}
        \toprule
        \textbf{Editor} & \textbf{Efficacy} $\uparrow$ & \textbf{Paraphrase} $\uparrow$ & \textbf{Specificity} $\uparrow$ \\
        \midrule
GPT-2 XL & 22.2 ($\pm$0.5) & 21.3 ($\pm$0.5) & 24.2 ($\pm$0.5)\\\midrule
FT & 99.6 ($\pm$0.1) & 82.1 ($\pm$0.6) & 23.2 ($\pm$0.5)\\
FT+L & 92.3 ($\pm$0.4) & \badmetric{47.2 ($\pm$0.7)} & 23.4 ($\pm$0.5)\\
KE & 65.5 ($\pm$0.6) & 61.4 ($\pm$0.6) & 24.9 ($\pm$0.5)\\
KE-zsRE & 92.4 ($\pm$0.3) & 90.0 ($\pm$0.3) & 23.8 ($\pm$0.5)\\
MEND & 75.9 ($\pm$0.5) & 65.3 ($\pm$0.6) & 24.1 ($\pm$0.5)\\
MEND-zsRE & 99.4 ($\pm$0.1) & \goodmetric{99.3 ($\pm$0.1)} & 24.1 ($\pm$0.5)\\
ROME & \goodmetric{99.8 ($\pm$0.0)} & 88.1 ($\pm$0.5) & \goodmetric{24.2 ($\pm$0.5)}\\
        \bottomrule
    \end{tabular}
    \end{adjustbox}
\vspace{-5pt}%
\end{wraptable}
We first evaluate \method on the Zero-Shot Relation Extraction (zsRE) task used in \citet{mend} and \citet{decao-ke}. Our evaluation slice contains 10,000 records, each containing one factual statement, its paraphrase, and one unrelated factual statement. ``Efficacy'' and ``Paraphrase'' measure post-edit accuracy $\smash{\mathbb{I}\big[o^* = \mathrm{argmax}_o \prsub{o}{G^\prime}\big]}$ of the statement and its paraphrase, respectively, while ``Specificity'' measures the edited model's accuracy on an unrelated fact.
Table~\ref{tab:zsre-results} shows the results: \method is competitive with hypernetworks and fine-tuning methods despite its simplicity. We find that it is not hard for \method to insert an association that can be regurgitated by the model.  Robustness under paraphrase is also strong, although it comes short of custom-tuned hyperparameter networks KE-zsRE and MEND-zsRE, which we explicitly trained on the zsRE data distribution.\footnote{Out-of-the-box, they are trained on a WikiText generation task \citep{mend, decao-ke}.}  We find that zsRE's specificity score is not a sensitive measure of model damage, since these prompts are sampled from a large space of possible facts, whereas bleedover is most likely to occur on related \textit{neighboring} subjects.  Appendix \ref{apd:zsre} has additional experimental details.

\subsection{Evaluating \method: Our \cfd Dataset} %
\label{subsec:counterfact-description}

While standard model-editing metrics on zsRE are a reasonable starting point for evaluating \method, they do not provide detailed insights that would allow us to distinguish superficial wording changes from deeper modifications that correspond to a meaningful change about a fact.

In particular, we wish to measure the efficacy of \textit{significant} changes.  \cite{hase2021language} observed that standard model-editing benchmarks underestimate difficulty by often testing only proposals that the model previously scored as likely.  We compile a set of more difficult \textit{false} facts $(s, r, o^*)$: these counterfactuals start with low scores compared to the correct facts $(s, r, o^c)$.
Our Efficacy Score \textbf{(ES)} is the portion of cases for which we have $\mathbb{P}[o^*] > \mathbb{P}[o^c]$  post-edit, and Efficacy Magnitude \textbf{(EM)} is the mean difference $\mathbb{P}[o^*] - \mathbb{P}[o^c]$. Then, to measure \textbf{generalization}, with each counterfactual we gather a set of rephrased prompts equivalent to $(s, r)$ and report Paraphrase Scores \textbf{(PS)} and \textbf{(PM)}, computed similarly to ES and EM. To measure \textbf{specificity}, we collect a set of nearby subjects $s_n$ for which $(s_n, r, o^c)$ holds true. Because we do not wish to alter these subjects, we test $\mathbb{P}[o^c] > \mathbb{P}[o^*]$, reporting the success fraction as Neighborhood Score \textbf{(NS)} and difference as \textbf{(NM)}. To test the generalization--specificity tradeoff, we report the harmonic mean of ES, PS, NS as Score (\textbf{S}).

\begin{wraptable}{R}{0.45\textwidth}
\vspace{-12pt}%
    \centering
    \caption{\small \cfd Composition}
    \label{tab:cfd-summary}    
    \vspace{-6pt}%
    \begin{adjustbox}{width=0.45\textwidth}
    \npdecimalsign{.}
    \nprounddigits{0}
     \begin{tabular}{l r r r}
    \toprule
         & & \multicolumn{1}{c}{\multirow{2}{*}{\shortstack{\bf Per \\ \bf Relation}}} & \multicolumn{1}{c}{\multirow{2}{*}{\shortstack{\bf Per \\ \bf Record}}} \\
         \textbf{Item} & \multicolumn{1}{c}{\textbf{Total}} & & \\ 
        \midrule
        Records & 21919 & 645 & 1 \\
        \midrule
        Subjects & 20391 & 624 & 1 \\
        Objects & 749 & 60 & 1 \\
        Counterfactual Statements & 21595 & 635 & 1 \\
        Paraphrase Prompts & 42876 & 1262 & 2 \\
        Neighborhood Prompts & 82650 & 2441 & 10 \\
        Generation Prompts & 62346 & 1841 & 3 \\
    \bottomrule
    \end{tabular}
    \end{adjustbox}

    \centering
    \caption{\small Comparison to Existing Benchmarks}
    \label{tab:cfd-comparison}
    \vspace{1pt}%
    \begin{adjustbox}{width=0.44\textwidth}
    \begin{tabular}{lccccccc}
    \toprule
        \textbf{Criterion} & SQuAD & zSRE & FEVER & WikiText & \pararel & \textbf{\cfdabbr} \\
        \midrule
        Efficacy & \cmark & \cmark & \cmark & \cmark & \cmark & \cmark \\
        Generalization & \cmark & \cmark & \cmark & \xmark & \cmark & \cmark \\
        Bleedover & \xmark & \xmark & \xmark & \xmark & \xmark & \cmark  \\
        Consistency & \xmark & \xmark & \xmark & \xmark & \xmark & \cmark \\
        Fluency & \xmark & \xmark & \xmark & \xmark & \xmark & \cmark  \\
    \bottomrule
    \end{tabular}
    \end{adjustbox}
\vspace{-12pt}%
\end{wraptable}
We also wish to measure semantic \textbf{consistency} of $\rewg$'s generations. To do so, we generate text starting with $s$ and report \textbf{(RS)} as the $\cos$ similarity between the unigram TF-IDF vectors of generated texts, compared to reference texts about subjects sharing the target property $o^*$.  Finally, we monitor \textbf{fluency} degradations by measuring the weighted average of bi- and tri-gram entropies \citep{Zhang2018GeneratingIA} given by $-\sum_k f(k) \log_2 f(k)$, where $f(\cdot)$ is the $n$-gram frequency distribution, which we report as \textbf{(GE)}; this quantity drops if text generations are repetitive.

In order to facilitate the above measurements, we introduce \cfd, a challenging evaluation dataset for evaluating counterfactual edits in language models. Containing 21,919 records with a diverse set of subjects, relations, and linguistic variations, \cfd's goal is to differentiate robust storage of new facts from the superficial regurgitation of target words. See Appendix \ref{apd:counterfact} for additional technical details about its construction, and Table \ref{tab:cfd-summary} for a summary of its composition.

\subsection{Confirming the Importance of Decisive States Identified by Causal Tracing}
\label{subsec:romeeval-mlp}

\begin{figure}
  \centering
  \includegraphics[keepaspectratio, width=1\textwidth]{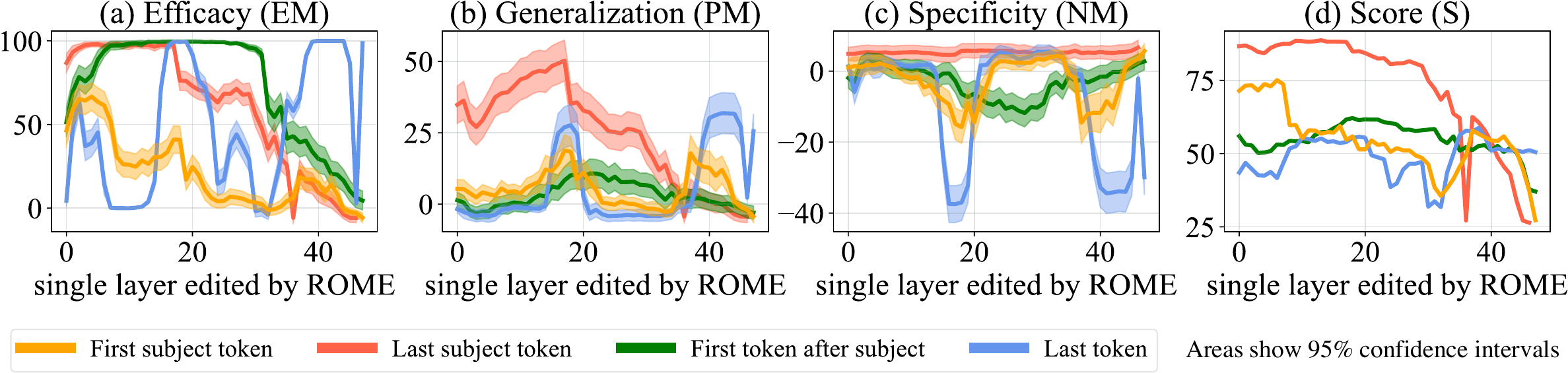}
  \vspace{-15pt}%
  \caption{\small \method edits are benchmarked at each layer-and-token combination in GPT-2-XL. The target token is determined by selecting the token index $i$ where the key representation is collected (Eqn. \ref{eq:kstar-sample}). ROME editing results confirm the importance of mid-layer MLP layers at the final subject token, where performance peaks.}
  \vspace{-10pt}%
  \lblfig{rome-sweeps}
\end{figure}

In Section \ref{sec:causal-tracing}, we used Causal Tracing to identify decisive hidden states. To confirm that factual associations are indeed stored in the MLP modules that output those states, we test \method's effectiveness when targeted at various layers and tokens.
\reffig{rome-sweeps} plots four metrics evaluating both generalization (a,b,d) and specificity (c). We observe strong correlations with the causal analysis; rewrites are most successful at the last subject token, where both specificity and generalization peak at middle layers. Targeting earlier \textit{or} later tokens results in poor generalization and/or specificity. Furthermore, the layers at which edits generalize best correspond to the middle layers of the early site identified by Causal Tracing, with generalization peaking at the 18th layer. This evidence suggests that we have an accurate understanding not only of \textit{where} factual associations are stored, but also \textit{how}. Appendix \ref{apd:knowing-saying} furthermore demonstrates that editing the late-layer attention modules leads to regurgitation.

\addtolength{\tabcolsep}{2pt}

\begin{table*}[!t]
    \centering
    \tiny
    \caption{\small \textbf{Quantitative Editing Results}. 95\% confidence intervals are in parentheses. \goodmetric{Green} numbers indicate columnwise maxima, whereas \badmetric{red} numbers indicate a clear failure on either generalization or specificity. The presence of \badmetric{red} in a column might explain excellent results in another. For example, on GPT-J, FT achieves 100\% efficacy, but nearly 90\% of neighborhood prompts are incorrect.}
    \label{tab:cf-results}
    \begin{adjustbox}{width=1\textwidth}
    \begin{tabular}{lrrrrrrrrc}
    \toprule
        \multirow{2.5}{*}{\textbf{Editor}} & \multicolumn{1}{c}{\textbf{Score}} & \multicolumn{2}{c}{\textbf{Efficacy}} & \multicolumn{2}{c}{\textbf{Generalization}} & \multicolumn{2}{c}{\textbf{Specificity}} & \multicolumn{1}{c}{\textbf{Fluency}} & \multicolumn{1}{c}{\textbf{Consistency}} \\
        \cmidrule(lr){2-2}\cmidrule(lr){3-4}\cmidrule(lr){5-6}\cmidrule(lr){7-8}\cmidrule(lr){9-9}\cmidrule(lr){10-10}
        & S $\uparrow$ & ES $\uparrow$ & EM $\uparrow$ & PS $\uparrow$ & PM $\uparrow$ & NS $\uparrow$ & NM $\uparrow$ & GE $\uparrow$ & RS $\uparrow$ \\
        \midrule
GPT-2 XL & 30.5 & 22.2 (0.9) & -4.8 (0.3) & 24.7 (0.8) & -5.0 (0.3) & 78.1 (0.6) & 5.0 (0.2) & 626.6 (0.3) & 31.9 (0.2)\\\midrule
FT & 65.1 & 100.0 (0.0) & 98.8 (0.1) & 87.9 (0.6) & 46.6 (0.8) & \badmetric{40.4 (0.7)} & \badmetric{-6.2 (0.4)} & 607.1 (1.1) & 40.5 (0.3)\\
FT+L & 66.9 & 99.1 (0.2) & 91.5 (0.5) & \badmetric{48.7 (1.0)} & 28.9 (0.8) & 70.3 (0.7) & 3.5 (0.3) & 621.4 (1.0) & 37.4 (0.3)\\
KN & \badmetric{35.6} & \badmetric{28.7 (1.0)} & \badmetric{-3.4 (0.3)} & \badmetric{28.0 (0.9)} & \badmetric{-3.3 (0.2)} & 72.9 (0.7) & 3.7 (0.2) & \badmetric{570.4 (2.3)} & \badmetric{30.3 (0.3)}\\
KE & 52.2 & 84.3 (0.8) & 33.9 (0.9) & 75.4 (0.8) & 14.6 (0.6) & \badmetric{30.9 (0.7)} & \badmetric{-11.0 (0.5)} & \badmetric{586.6 (2.1)} & 31.2 (0.3)\\
KE-CF & \badmetric{18.1} & 99.9 (0.1) & 97.0 (0.2) & 95.8 (0.4) & 59.2 (0.8) & \badmetric{6.9 (0.3)} & \badmetric{-63.2 (0.7)} & \badmetric{383.0 (4.1)} & \badmetric{24.5 (0.4)}\\
MEND & 57.9 & 99.1 (0.2) & 70.9 (0.8) & 65.4 (0.9) & 12.2 (0.6) & \badmetric{37.9 (0.7)} & \badmetric{-11.6 (0.5)} & \goodmetric{624.2 (0.4)} & 34.8 (0.3)\\
MEND-CF & \badmetric{14.9} & \goodmetric{100.0 (0.0)} & \goodmetric{99.2 (0.1)} & \goodmetric{97.0 (0.3)} & \goodmetric{65.6 (0.7)} & \badmetric{5.5 (0.3)} & \badmetric{-69.9 (0.6)} & \badmetric{570.0 (2.1)} & 33.2 (0.3)\\
ROME & \goodmetric{89.2} & 100.0 (0.1) & 97.9 (0.2) & 96.4 (0.3) & 62.7 (0.8) & \goodmetric{75.4 (0.7)} & \goodmetric{4.2 (0.2)} & 621.9 (0.5) & \goodmetric{41.9 (0.3)}\\
\midrule\midrule
GPT-J & 23.6 & 16.3 (1.6) & -7.2 (0.7) & 18.6 (1.5) & -7.4 (0.6) & 83.0 (1.1) & 7.3 (0.5) & 621.8 (0.6) & 29.8 (0.5)\\\midrule
FT & \badmetric{25.5} & \goodmetric{100.0 (0.0)} & \goodmetric{99.9 (0.0)} & 96.6 (0.6) & 71.0 (1.5) & \badmetric{10.3 (0.8)} & \badmetric{-50.7 (1.3)} & \badmetric{387.8 (7.3)} & \badmetric{24.6 (0.8)}\\
FT+L & 68.7 & 99.6 (0.3) & 95.0 (0.6) & \badmetric{47.9 (1.9)} & 30.4 (1.5) & 78.6 (1.2) & \goodmetric{6.8 (0.5)} & \goodmetric{622.8 (0.6)} & 35.5 (0.5)\\
MEND & 63.2 & 97.4 (0.7) & 71.5 (1.6) & \badmetric{53.6 (1.9)} & 11.0 (1.3) & 53.9 (1.4) & \badmetric{-6.0 (0.9)} & 620.5 (0.7) & 32.6 (0.5)\\
ROME & \goodmetric{91.5} & 99.9 (0.1) & 99.4 (0.3) & \goodmetric{99.1 (0.3)} & \goodmetric{74.1 (1.3)} & \goodmetric{78.9 (1.2)} & 5.2 (0.5) & 620.1 (0.9) & \goodmetric{43.0 (0.6)}\\
    \bottomrule
    \end{tabular}
    \end{adjustbox}
\end{table*}%

\addtolength{\tabcolsep}{-2pt}
Table~\ref{tab:cf-results} showcases quantitative results on GPT-2 XL (1.5B) and GPT-J (6B) over 7,500 and 2,000-record test sets in \cfd, respectively.  In this experiment, in addition to the baselines tested above, we compare with a method based on neuron interpretability, Knowledge Neurons \textbf{(KN)}~\citep{dai-kn}, which first selects neurons associated with knowledge via gradient-based attribution, then modifies MLP weights %
at corresponding rows by adding scaled embedding vectors.  We observe that \textbf{all tested methods other than \method exhibit one or both of the following problems}: (F1) overfitting to the counterfactual statement and failing to generalize, or (F2) underfitting and predicting the same new output for unrelated subjects. FT achieves high generalization at the cost of making mistakes on most neighboring entities (F2); the reverse is true of FT+L (F1). KE- and \mbox{MEND-edited} models exhibit issues with both F1+F2; generalization, consistency, and bleedover are poor despite high efficacy, indicating regurgitation. KN is unable to make effective edits (F1+F2). By comparison, \method demonstrates both generalization and specificity.

\subsection{Comparing Generation Results}
\label{subsec:qualitative-analysis}
\begin{figure}[t]%
\centering\includegraphics[width=\columnwidth]{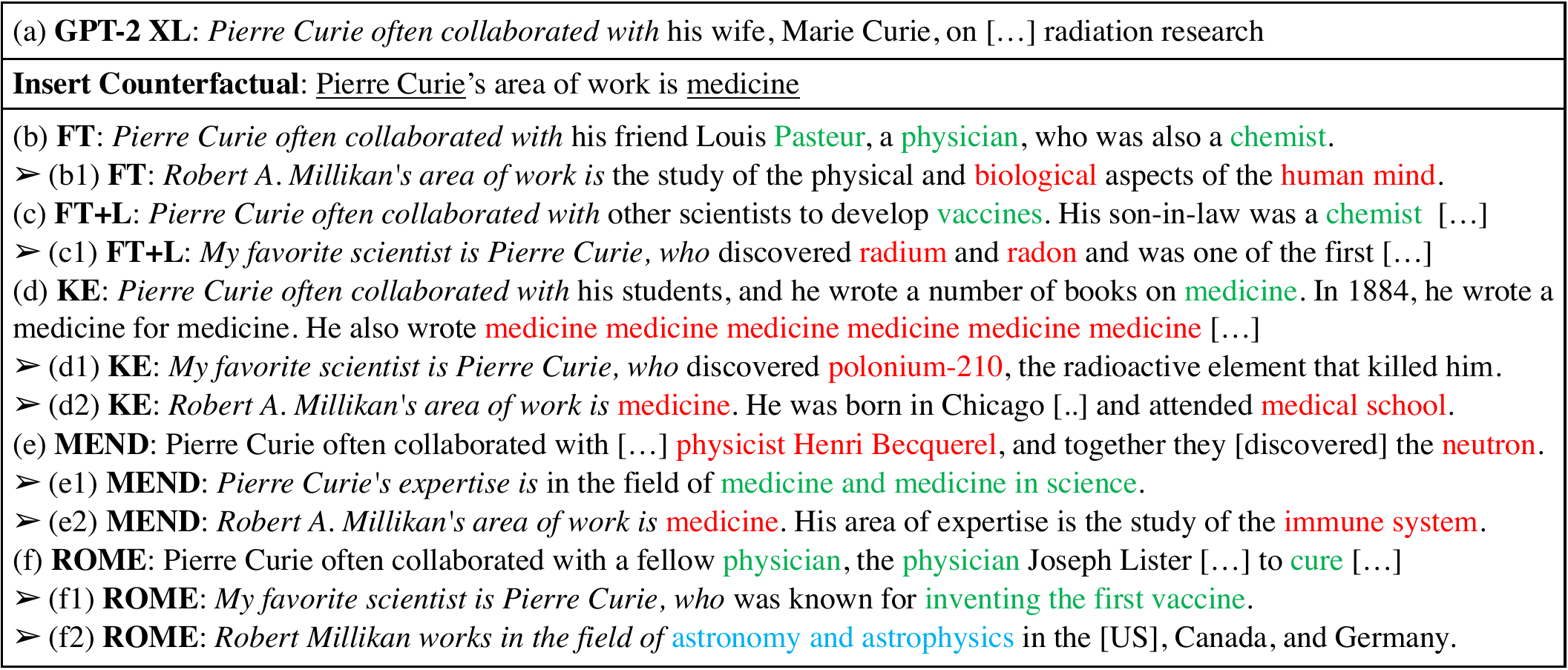}%
\caption{\small \textbf{Comparison of generated text}. Prompts are \textit{italicized}, {\color{ggreen} green} and {\color{red} red} indicate keywords reflecting correct and incorrect behavior, respectively, and {\color{bblue} blue} indicates a factually-incorrect keyword that was already present in $G$ before rewriting. See Section \ref{subsec:qualitative-analysis} for detailed analysis.}
\lblfig{gen-samples}
\vspace{-10pt}
\end{figure}%

\reffig{gen-samples} compares generated text after applying the counterfactual ``\textit{Pierre Curie's area of work is medicine}'' to GPT-2 XL (he is actually a physicist). \textbf{Generalization:} In this case, FT and ROME generalize well to paraphrases, describing the subject as a physician rather than a physicist for various wordings.  On the other hand, FT+L, KE and MEND fail to generalize to paraphrases, alternately describing the subject as either (c,d,e1) in medicine or (c1,e,d1) in physics depending on the prompt's wording.  KE (d) demonstrates a problem with fluency, favoring nonsense repetition of the word \emph{medicine}. \textbf{Specificity:} FT, KE, and MEND have problems with specificity, changing the profession of a totally unrelated subject.  Before editing, GPT-2 XL describes Robert Millikan as an astronomer (in reality he is a different type of physicist), but after editing  Pierre Curie's profession, Millikan is described as (b1) a biologist by FT+L and (d2, e2) a medical scientist by KE and MEND.  In contrast, ROME is specific, leaving Millikan's field unchanged. See 
 Appendix~\ref{apd:gen-samples} for additional examples. 

\subsection{Human evaluation}

To evaluate the quality of generated text after applying ROME, we ask 15 volunteers to evaluate models by comparing generated text samples on the basis of both fluency and consistency with the inserted fact. Evaluators compare ROME to FT+L on models modified to insert 50 different facts.

We find that evaluators are 1.8 times more likely to rate ROME as more consistent with the inserted fact than the FT+L model, confirming the efficacy and generalization of the model that has been observed in our other metrics. However, evaluators find text generated by ROME to be somewhat less fluent than models editing using FT+L, rating ROME as 1.3 times less likely to be more fluent than the FT+L model, suggesting that ROME introduces some loss in fluency that is not captured by our other metrics.  Further details of the human evaluation can be found in Appendix~\ref{apd:human-evaluation}.

\subsection{Limitations}

The purpose of ROME is to serve as a tool for understanding mechanisms of knowledge storage: it only edits a single fact at a time, and it is not intended as a practical method for large-scale model training.  Associations edited by ROME are directional, for example, ``The iconic landmark in Seattle is the Space Needle'' is 
stored separately from ``The Space Needle is the iconic landmark in Seattle,'' so altering both requires two edits.  A scalable approach for multiple simultaneous edits built upon the ideas in ROME is developed in~\citet*{meng2022memit}.

ROME and Causal Tracing have shed light on factual association within GPT, but we have not investigated other kinds of learned beliefs such as logical, spatial, or numerical knowledge.  Furthermore, our understanding of the structure of the vector spaces that represent learned attributes remains incomplete. Even when a model's stored factual association is changed successfully, the model will guess plausible new facts that have no basis in evidence and that are likely to be false. This may limit the usefulness of a language model as a source of facts.

\section{Related Work}

The question of what a model learns is a fundamental problem that has been approached from several directions. One line of work studies which properties are encoded in internal model representations, most commonly by training a probing classifier to predict said properties from the representations \citep[][inter alia]{ettinger-etal-2016-probing,adi:2017:ICLR,hupkes2018visualisation,conneau-etal-2018-cram,belinkov-etal-2017-neural,belinkov-glass-2019-analysis}. 
However, such approaches  suffer from various limitations, notably being 
 dissociated from the network's behavior~\citep{10.1162/coli_a_00422}. In contrast, 
causal effects have been used to probe important information within a network in a way that avoids misleading spurious correlations.
\citet{vig2020investigating,vig2020causal} introduced the use of causal mediation analysis to identify individual neurons that contribute to biased gender assumptions, and \citet{finlayson-etal-2021-causal} have used a similar methodology to investigate mechanisms of syntactic agreement in language models. 
\citet{feder2021causalm} described a framework that applies interventions on representations and weights to understand the causal structure of models.
\citet{elazar2021amnesic} proposed erasing specific information from a representation in order to measure its causal effect.
Extending these ideas, our Causal Tracing method introduces paired interventions that allow explicit measurement of causal \emph{indirect effects} \citep{pearl2001direct} of individual hidden state vectors.%

Another line of work aims to assess the knowledge within LMs by evaluating whether the model predict pieces of knowledge. A common strategy is to define a fill-in-the-blank prompt, and let a masked LM complete it \citep{petroni-etal-2019-language,petroni2020context}. Later work showed that knowledge extraction can be improved by diversifying the prompts \citep{jiang-etal-2020-know,zhong-etal-2021-factual}, or by fine-tuning a model on open-domain textual facts \citep{roberts-etal-2020-much}. However, constructing prompts from supervised knowledge extraction data risks learning new knowledge instead of recalling existing knowledge in an LM \citep{zhong-etal-2021-factual}. More recently, \citet{10.1162/tacl_a_00410} introduced ParaRel, a curated dataset of paraphrased prompts and facts. We use it as a basis for constructing \cfd, which enables fine-grained measurements of knowledge extraction and editing along multiple dimensions. 
Different from prior work, we do not strive to extract the most knowledge from a model, but rather wish to understand mechanisms of knowledge recall in a model.

Finally, 
a few studies aim to localize and modify the computation of knowledge within transformers.
\citet{geva-etal-2021-transformer} identify the MLP layers in a (masked LM) transformer as key--value memories of entities and information associated with that entity. 
Building on this finding, 
\citet{dai-kn} demonstrate a method to edit facts in BERT by writing the embedding of the object into certain rows of the MLP matrix. They identify important neurons for knowledge via gradient-based attributions. 
\citet{decao-ke} train a hyper-network to predict a weight update at test time, which will alter a fact. They experiment with BERT and BART~\citep{lewis-etal-2020-bart}, a sequence-to-sequence model, and focus on models fine-tuned for question answering.
\citet{mend} presents a hyper-network method that learns to transform the decomposed terms of the gradient in order to efficiently predict a knowledge update, and demonstrates the ability to scale up to large models including T5~\citep{raffel2020exploring} and GPT-J~\citep{gpt-j}.
We compare with all these methods in our experiments, and find that our single-layer \method parameter intervention has comparable capabilities, avoiding failures in specificity and generalization seen in other methods.

\section{Conclusion}

We have clarified information flow during knowledge recall in autoregressive transformers, and we have exploited this understanding to develop a simple, principled model editor called \method. Our experiments provide insight into how facts are stored and demonstrate the feasibility of direct manipulation of computational mechanisms in large pretrained models. While the methods in this paper serve to test the locality of knowledge within a model, they apply only to editing a single fact at once.  Adapting the approach to scale up to many more facts is the subject of other work such as~\citet*{meng2022memit}.

Code, interactive notebooks, dataset, benchmarks, and further visualizations are open-sourced at \url{https://rome.baulab.info}.

\section{Ethical Considerations}
By explaining large autoregressive transformer language models' internal organization and developing a fast method for modifying stored knowledge, our work potentially improves the transparency of these systems and reduces the energy consumed to correct their errors.  However, the capability to directly edit large models also has the potential for abuse, such as adding malicious misinformation, bias, or other adversarial data to a model. Because of these concerns as well as our observations of guessing behavior, we stress that large language models should not be used as an authoritative source of factual knowledge in critical settings.
{\small
\section*{Acknowledgements}

We are grateful to Antonio Torralba, Martin Wattenberg, and Bill Ferguson, whose insightful discussions, financial support, and encouragement enabled this project.  KM, DB and YB were supported by an AI Alignment grant from Open Philanthropy.  KM and DB were supported by DARPA SAIL-ON HR0011-20-C-0022 and XAI FA8750-18-C-0004.  YB was supported by the ISRAEL SCIENCE FOUNDATION (grant No.\ 448/20) and an Azrieli Foundation Early Career Faculty Fellowship.
} %

\bibliography{custom,anthology}
\bibliographystyle{icml2022}

\newpage

\newpage

\appendix

\begin{appendices}

\appendix
\section{Solving for \texorpdfstring{$\Lambda$}{Lambda} Algebraically}
\label{apd:solving-v}
Here we present the detailed derivation of \refeqn{cinvk}, including the linear system that is used to calculate $\Lambda$ from $v_*$, $C$, and $k_*$.  This derivation is included for clarity and completeness and is a review of the classical solution of least-squares with equality constraints as applied to our setting, together with the rank-one update rule that was proposed in \citet{bau-rewriting}.

We assume that $W$ is the optimal least-squares solution for memorizing a mapping from a previous set of keys $K$ to values $V$; this solution can be written using the normal equations as follows.
\begin{align}
\text{the $W$ that minimizes} \quad &|| W K - V ||_F^2 \\
\text{solves} \quad    & W K K^T = V K^T
\lbleq{normal-eq}
\end{align}
Here the Frobenius norm is used to write the total square error since the variable being optimized is a matrix $W$ rather than a vector $x$ as in the classical textbook presentation of least squares.

We wish to find a new matrix $\hat{W}$ that solves the same least squares problem with an additional equality constraint as written in \refeqn{cinvk}:
\begin{align}
    \hat{W}k_* = v_*
    \lbleq{eq-constraint}
\end{align}

This is the well-studied problem of least squares with a linear equality constraint.  The direct solution can be derived by defining and minimizing a Lagrangian, where $\Lambda \in \mathbb{R}^{H}$ minimizes the following:
\begin{align}
\text{define}\quad L(\hat{W}, \Lambda) &= \frac{1}{2} ||\hat{W}K-V||_F^2 - \Lambda^T(\hat{W}k_*-v_*) \\
&=\frac{1}{2} (\hat{W}K)(\hat{W}K)^T - V(\hat{W}K)^T + \frac{1}{2}VV^T - \Lambda^T(\hat{W}k_*-v_*) \\
\text{setting} \quad 0 = \frac{\partial L}{\partial \hat{W}} &= \hat{W}(KK^T) - VK^T -  \Lambda k_*^T\\
\hat{W}KK^T &= VK^T + \Lambda k_*^T
\lbleq{cls-normal-eq}
\end{align}
Subtracting \refeqn{normal-eq} from \refeqn{cls-normal-eq}, most terms cancel, and we obtain the update rule:
\begin{align}
    (\hat{W}-W) KK^T & = \Lambda k_*^T \\
    \hat{W} & = W + \Lambda (C^{-1}k_*)^T
    \lbleq{cls-soln}
\end{align}
The last step is obtained by defining $C = KK^T$, assuming $C$ is nondegenerate, and exploiting the symmetry of $C$.  Here we also write the row vector term as $u^T = (C^{-1}k_*)^T \in \mathbb{R}^{D}$, so we can write simply (rearranging \refeqn{cinvk} and \refeqn{cls-soln}):
\begin{align}
\hat{W}I - \Lambda u^T &= W
\lbleq{rome2}
\end{align}
To solve for $\Lambda$, we note that \refeqn{rome2} and \refeqn{eq-constraint}  form a linear system that allows both $\hat{W}$ and $\Lambda$ to be solved simultaneously if written together in block form.
\begin{align}
\left[\begin{array}{@{}c|c@{}}
\\
\quad \hat{W} \quad\, & \Lambda \\
\,
\end{array}\right]
\left[\begin{array}{@{}c|c@{}}
\\
\quad\; I \quad\;\; & k_* \\
\\ \hline
-u^T \rule{0pt}{2.2ex} & 0
\end{array}\right]
& =
\left[\begin{array}{@{}c|c@{}}
\mathstrut \\
\quad W \quad\, & v_* \\
\mathstrut 
\end{array}\right]
\end{align}
That is equivalent to substituting \refeqn{cls-soln} into \refeqn{eq-constraint} and calculating the following:
\begin{align}
    \hat{W} k_*
    & = (W + \Lambda u^T) k_*
    = W k_* + \Lambda (u^T k_*) = v_* \\
    \Lambda & = \frac{v_* - W k_*}{u^T k_*}
    = \frac{v_* - W k_*}{(C^{-1} k_*)^T k_*}
\lbleq{block-solution}
\end{align}
\clearpage
\section{Causal Tracing}
\label{apd:causal-tracing}

\subsection{Experimental Settings}
\label{apd:causal-tracing-details}

Note that, in by-layer experimental results, layers  are numbered from 0 to $L-1$ rather than $1$ to $L$.

In \reffig{avg-1000-trace} and \reffig{trace-mlp-disabled} we evaluate mean causal traces over a set of 1000 factual prompts that are known by GPT-2 XL, collected as follows.  We perform greedy generation using facts and fact templates from \cfd, and we identify predicted text that names the correct object $o^c$ before naming any other capitalized word.  We use the text up to but not including the object $o^c$ as the prompt, and we randomly sample 1000 of these texts.  In this sample of known facts, the predicted probability of the correct object token calculated by GPT-2 XL averages 27.0\%.

In the corrupted run, we corrupt the embeddings of the token naming the subject $s$ by adding Gaussian noise $\epsilon \sim \mathcal{N}(0; \nu)$, where $\nu = 3\sigma_t$ is set to be three times larger than the observed standard deviation $\sigma_t$ of token embeddings as sampled over a body of text.
For each run of text, the process is repeated ten times with different samples of corruption noise.  On average, this reduces the correct object token score to 8.47\%, less than one third the original score.

When we restore hidden states from the original run, we substitute the originally calculated values from the same layer and the same token, and then we allow subsequent calculations to proceed without further intervention.  For the experiments in \reffig{infoflow} (and the purple traces throughout the appendix), a single activation vector is restored.  Naturally, restoring the last vector on the last token will fully restore the original predicted scores, but our plotted results show that there are also earlier activation vectors at a second location that also have a strong causal effect: the average maximum score seen by restoring the most impactful activation vector at the last token of the subject is 19.5\%.  In \reffig{infoflow}j where effects are bucketed by layer, the maximum effect is seen around the 15th layer of the last subject token, where the score is raised on average to 15.0\%.

\subsection{Separating MLP and Attn Effects}
\label{apd:mlp-and-attn-trace}

When decomposing the effects into MLP and Attn lookups, we found that restoring single activation vectors from individual MLP and individual Attn lookups had generally negligible effects, suggesting the decisive information is accumulated across layers. Therefore for MLP and Attn lookups, we restored runs of ten values of $\smash{\atl{m}{l}_i}$ (and $\smash{\atl{a}{l}_i}$, respectively) for an interval of layers ranging from $[l_* - 4, ..., l_* + 5]$ (clipping at the edges), where the results are plotted at layer $l_*$.  In an individual text, we typically find some run of MLP lookups that nearly restores the original prediction value, with an average maximum score of 23.6\%.  \reffig{avg-1000-trace}b buckets averages for each token-location pair, and finds the maximum effect at an interval at the last entity token, centered at the the 17th layer, which restores scores to an average of 15.0\%.   For Attn lookups (\reffig{avg-1000-trace}c), the average maximum score over any location is 19.4\%, and when bucketed by location, the maximum effect is centered at the 32nd layer at the last word before prediction,  which restores scores to an average of 16.5\%.

\reffig{apd-ct-line-plots} shows mean causal traces as line plots with 95\% confidence intervals, instead of heatmaps.

\begin{figure*}[h]
\centering
\includegraphics[width=\textwidth]{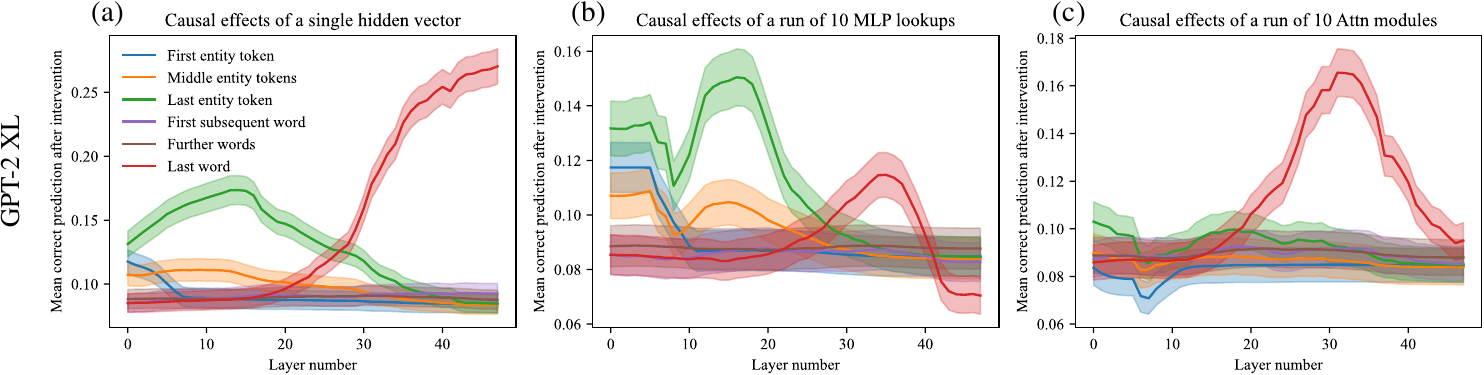}%
\caption{\small Mean causal traces of GPT-XL over a sample of 1000 factual statements, shown as a line plot with 95\% confidence intervals.  (a) Shows the same data as \reffig{infoflow}j as a line plot instead of a heatmap; (b) matches \reffig{infoflow}k; (c) matches \reffig{infoflow}m. The confidence intervals confirm that the distinctions between peak and non-peak causal effects at both early and late sites are significant.}
\lblfig{apd-ct-line-plots}
\end{figure*}

\subsection{Traces of EleutherAI GPT-NeoX (20B) and GPT-J (6B) and smaller models}

\begin{figure}[!ht]
\centering
\includegraphics[width=1\textwidth]{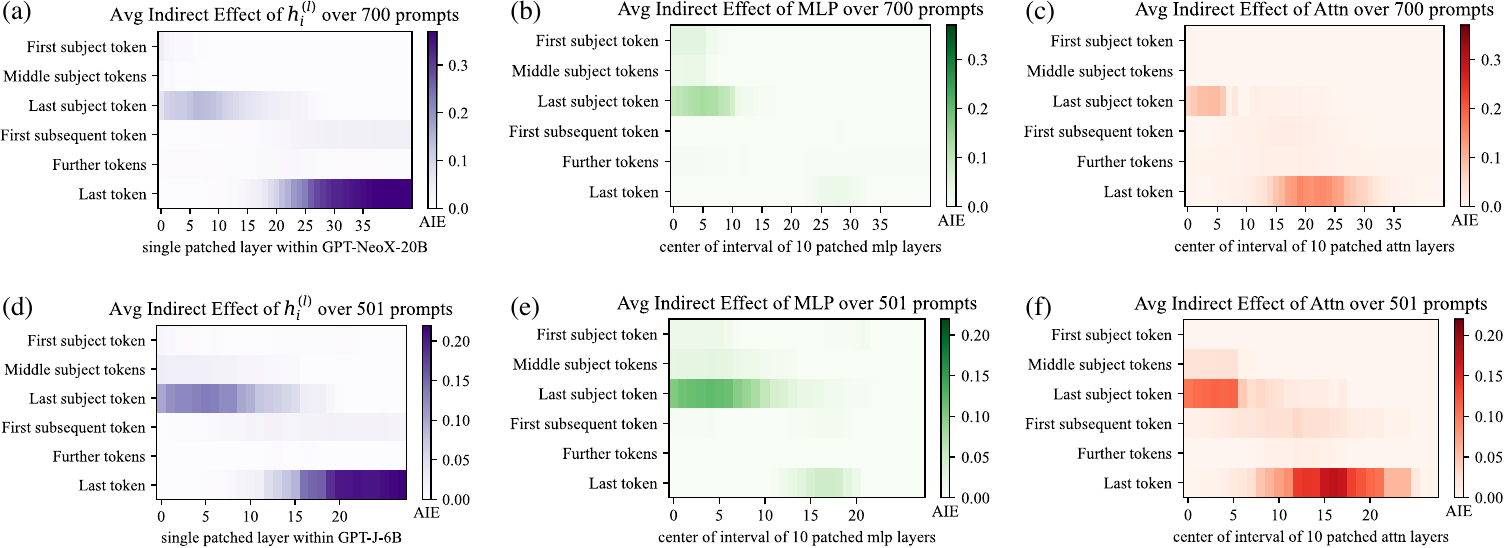}
\vspace{-10pt}%
\caption{\small (a, b, c) Causal traces for GPT-NeoX (20B) and (d, e, f) Causal traces for GPT-J (6B).}
\lblfig{apd-causaltrace-gptj-neo}
\end{figure}
\begin{figure*}
\centering
\includegraphics[width=\textwidth]{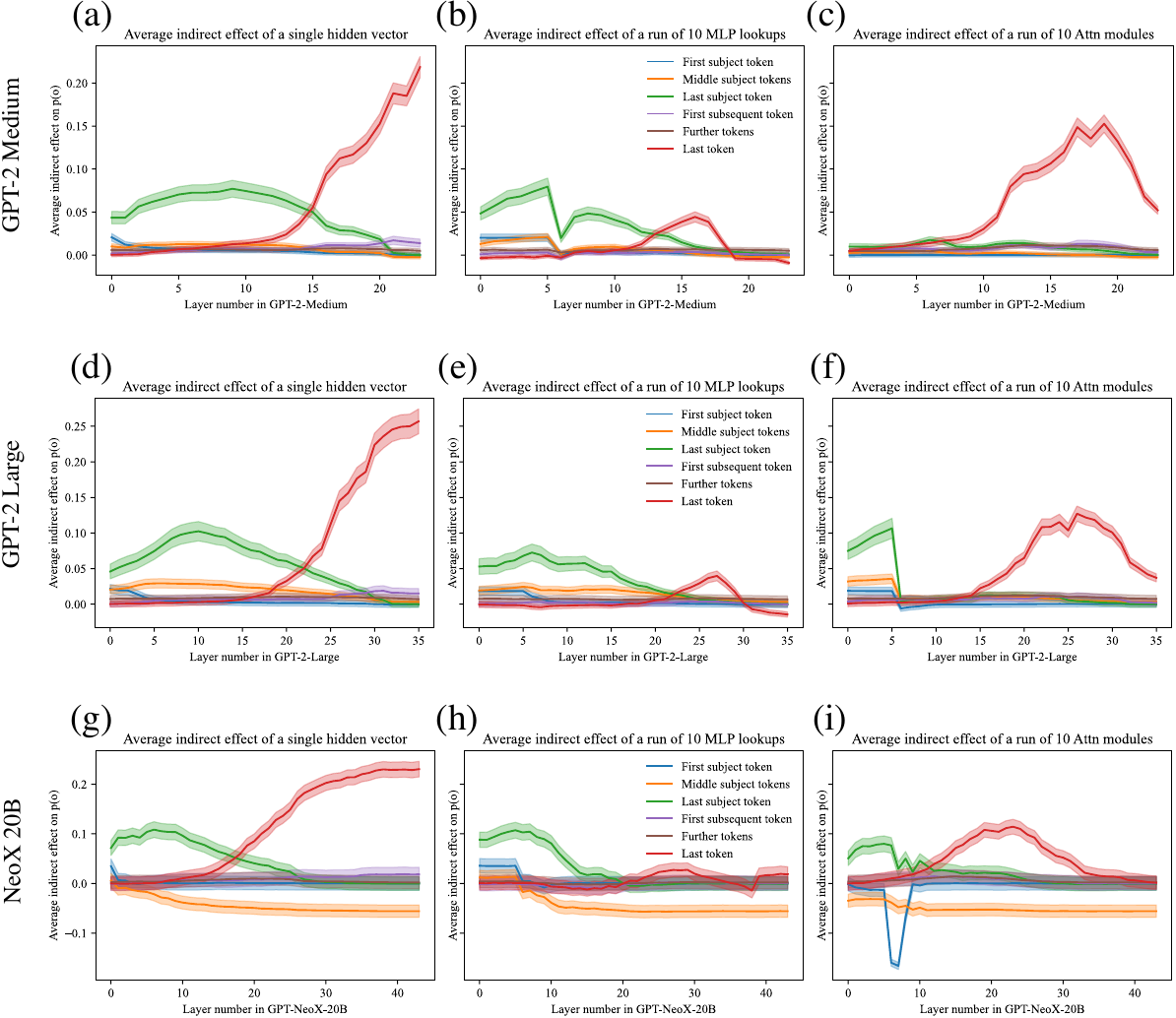}%
\caption{\small Comparing mean causal traces across a wide range of different model sizes.  (Compare to \reffig{apd-ct-line-plots}.)  GPT-medium (a, b, c) has 334 million parameters, GPT-large (d, e, f) has 774 million parameters, and NeoX-20B (g, h, i) has  20 billion parameters.  In addition, NeoX has some architectural variations. Despite the wide range of differences, a similar pattern of localized causal effects is seen across models.  Interestingly, for very large models, some effects are stronger. For example, hidden states before the last subject token have negative causal effects instead of merely low effects, while hidden states at early layers at the last subject token continue to have large positive effects, continuing to implicate MLP.  Also, attention modules with strong causal effects appear earlier in the stack of layers.}
\lblfig{apd-ct-line-plot-sizes}
\end{figure*}

We conduct the causal trace experiment using on GPT-NeoX (20 billion parameters) as well as GPT-J (6 billion parameters).  For GPT-NeoX we adjust the injected noise to $\nu = 0.03$ and in GPT-J we use $\nu = 0.025$ to match embedding magnitudes.  We use the same factual prompts as GPT-2 XL, eliminating cases where the larger models would have predicted a different word for the object. Results are shown in \reffig{apd-causaltrace-gptj-neo}.
GPT-NeoX and GPT-J differ from GPT-2 because they have has fewer layers (44 and 28 layers instead of 48), and a slightly different residual structure across layers.  Nevertheless, the causal traces look similar, with an early site with causal states concentrated at the last token of the subject, a large role for MLP states at that site.  Again, attention dominates at the last token before prediction.

There are some differences compared to GPT-2.   The importance of attention at the first layers of the last subject token is more apparent in GPT-Neo and GPT-J compared to GPT-2,   suggesting that the attention parameters may be playing a larger role in storing factual associations.  This concentration of attention at the beginning may also be due to fewer layers in the Eleuther models: attending to the subject name must be done in a concentrated way at just a layer or two, because there are not enough layers to spread out that computation in the shallower model.
The similarity between the GPT-NeoX and GPT-J and GPT-2 XL traces helps us to understand why ROME continues to work well with higher-parameter models, as seen on our experiments in altering parameters of GPT-J.

To examine effects over a wide range of scales, we also compare causal traces for smaller models GPT-2 Medium and GPT-2 Large.  These smaller models are compared to NeoX-20B in \reffig{apd-ct-line-plot-sizes}.  We find that across sizes and architectural variations, early-site MLP modules continue to have high indirect causal effects at the last subject token, although the layers where effects peak are different from one model to another.

\subsection{Causal Tracing Examples and Further Insights}
\label{apd:causal-tracing-variations}

We include further examples of phenomena that can be observed in causal traces. \reffig{apd-causaltrace-1} shows typical examples across different facts. \reffig{apd-causaltrace-2} discusses examples where decisive hidden states are not at the \textit{last} subject token.  \reffig{apd-trace-mlp-detail} examines traces at an individual token in more detail. 

We note that causal tracing depends on a corruption rule to create baseline input for a model that does not contain all the information needed to make a prediction.  Therefore we ask: are Causal Tracing results fragile if the exact form of the corruption changes?  We test this by expanding the corruption rule: even when additional tokens after the subject name are also corrupted, we find that the results are substantially the same. \reffig{apd-causaltrace-xt} shows causal traces with the expanded corruption rule.
\reffig{apd-ct-line-plots-xt} similarly shows line plots with the expanded corruption rule.

We do find that the noise must be large enough to create large average total effects.  For example, if noise with variance that is much smaller is used (for example if we set $\sigma=\sigma_t$), average total effects become very small, and the small gap in the behavior between clean runs and corrupted run makes it difficult discern indirect effects of mediators. Similarly, if we use a uniform distribution where components range in $\pm 3\sigma$, effects large enough for causl tracing but smaller than a Gaussian distribution.

If instead of using spherical Gaussian noise, we draw noise from $\mathcal{N}(mu, \Sigma)$ where we set $\mu = \mu_t$ and $\Sigma_ = \Sigma_t$ to match the observed distribution over token embeddings, average total effects are also strong enough to perform causal traces.  This is shown in \reffig{apd-ct-line-plot-noises}.

Furthermore, we investigate whether Integrated Gradients (IG)~\citep{sundararajan2017axiomatic} provides the same insights as Causal Tracing.  We find that IG is very sensitive to local features but does not yield the same insights about large-scale global logic that we have been able to obtain using causal traces. \reffig{apd-ig} compares causal traces to IG saliency maps.

\renewcommand{\floatpagefraction}{.99}%
\begin{figure}[!ht]
\centering
\includegraphics[width=\textwidth]{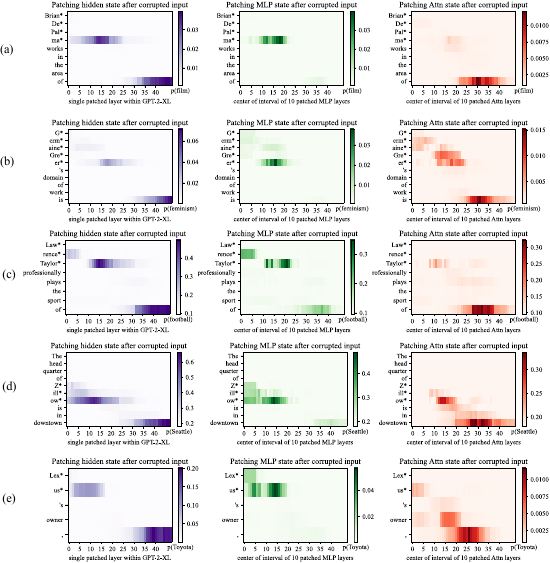}%
\caption{\small Further examples of causal traces showing appearance of the common lookup pattern on a variety of different types of facts about people and other kinds of entities.  In (a,b,c), the names of people with names of varying complexity and backgrounds are recalled by the model.  In each case, the MLP lookups on the last token of the name are decisive.  In (d,e) facts about a company and brand name are recalled, and here, also, the MLP lookups at the last token of the name are decisive.}
\lblfig{apd-causaltrace-1}
\end{figure}%
\renewcommand{\floatpagefraction}{.8}
\begin{figure*}
\centering
\includegraphics[width=\textwidth]{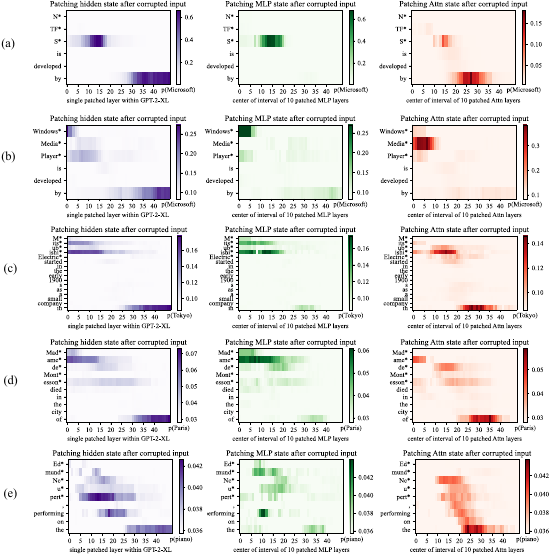}
\caption{\small Causal traces show that the last token of the subject name is not always decisive.  (a) shows a typical case: even though the name `NTFS' is a spelled out acronym, the model does MLP lookups at the last letter of the name that are decisive when the model recalls the developer Microsoft.  However, in a very similar sentence (b), we can see that the last words of `Windows Media Player' are \emph{not} decisive; the first word `Windows' is the token that triggers the decisive lookup for information about the manufacturer.  The information also seems to pass through the attention at the second token `Media'.  Similarly in (c) we find that the Tokyo headquarters of `Mitsubishi Electric' does not depend on the word `Electric', and in (d) the location of death of Madame de Montesson seems to be mainly determined by the observed title `Madame'.  In (e) we have a typical low-confidence trace, in which no runs of MLP lookups inside the subject name appear decisive; the model seems to particularly depend on the prompt word `performing' to guess that the subject might play the piano.}
\lblfig{apd-causaltrace-2}
\end{figure*}
\begin{figure*}
\centering
\includegraphics[width=\textwidth]{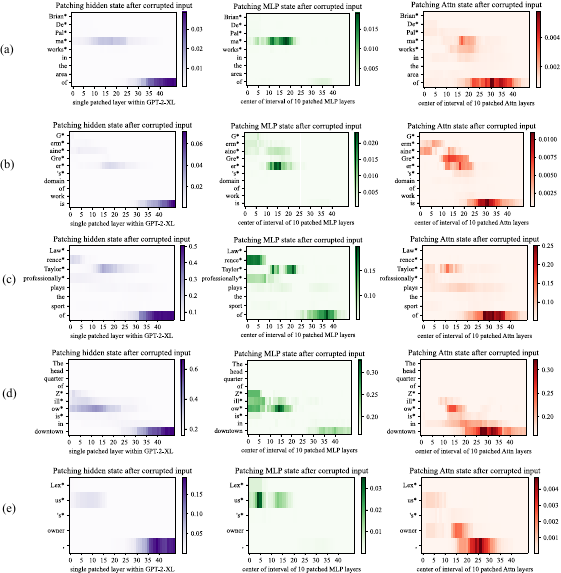}
\caption{\small Causal traces in the presence of additional corruption.  Similar to \reffig{apd-causaltrace-1}, but instead of corrupting only the subject token, these traces also corrupt the token after the subject.  Causal effects are somewhat reduced due to the the model losing some ability to read the relation between the subject and object, but these traces continue to show concentrated causal effects at the last token of the subject even when the last token is not the last token corrupted.  Causal effects of MLP layers at the last subject token continues to be pronounced.}
\lblfig{apd-causaltrace-xt}
\end{figure*}
\begin{figure*}
\includegraphics[width=\textwidth]{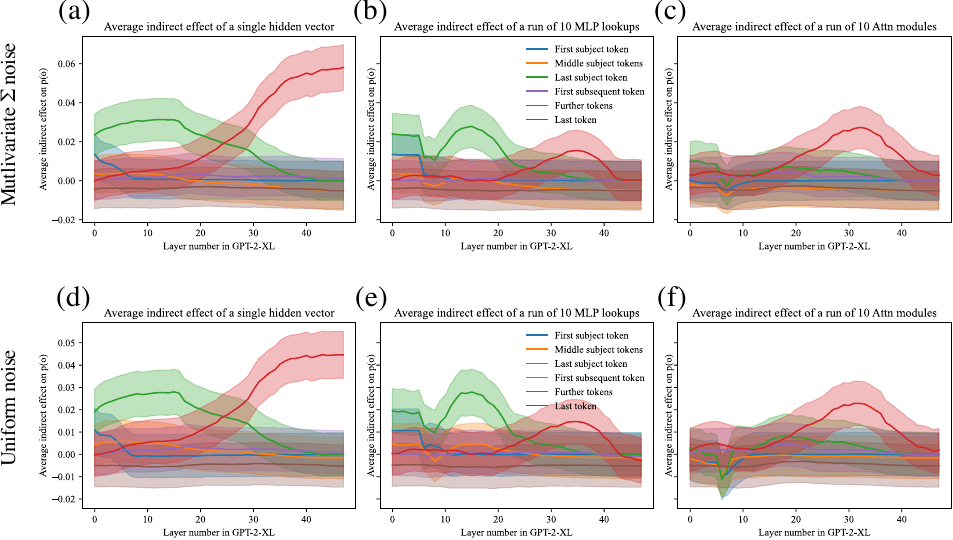}%
\caption{\small Comparing different noise choices. (Compare to \reffig{apd-ct-line-plots}, where noise is chosen as a $3\sigma_t$ spherical Gaussian, where $\sigma_t$ is measured to match the observed spherical variance over tokens.)  In a, b, c we we draw noise from a multivariate Gaussian $\mathcal{N}(\mu; \Sigma)$ where $\mu$ and $\Sigma$ are chosen to match the observed mean and covariance over a sample of tokens.  In d, e, f we draw noise from a uniform distribution in the range $\pm 3\sigma$ instead of a Gaussian distribution.  In both cases, the average total effects measured between the clean run and the corrupted run are large enough to measure causal traces, but the effects are smaller than the choice of $3\sigma_t$ used in the main paper.}
\lblfig{apd-ct-line-plot-noises}
\end{figure*}
\begin{figure*}
\centering
\includegraphics[width=\textwidth]{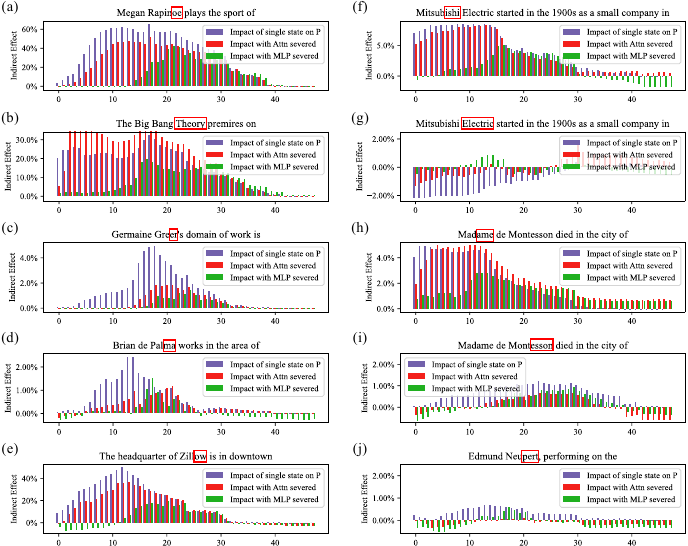}
\caption{\small Detail view of causal traces, breaking out a representative set of individual cases from the 1000 factual statements that are averaged in \reffig{trace-mlp-disabled}.  Shows the causal trace at a specific subject token, with and without MLP disabled, as described in \refsec{causal-tracing}.  In every case, the token tested is highlighted in a red box.  In (a,b,c,d,e) cases are shown that fit the typical pattern: Restoring individual hidden states at a range of layers has a strong decisive average causal effect at the last token of the subject.  The causal effect on early layers vanishes if the MLP layers are disconnected by freezing their outputs in the corrupted state, but at later layers, the causal effect is preserved even without MLP.   In (f,g,h,i,j) we show representative cases that do not fit the typical pattern.  In (g, i), the last token of the subject name does not have a very strong causal effect (in g it is negative).  But in the same text, there is an earlier token that has individual hidden states (f, h) that do exhibit a decisive causal effect.  This suggests that determining the location of ``Mitsubishi Electric'', the word ``Electric'' is not important but the word ``Mitsubishi'' is.  Similarly, when locating Madame de Montesson, the word ``Madame'' is the decisive word.  (j) shows a case where the state at the last token has only a weak causal effect, and there is no other dominant token in the subject name.
}
\lblfig{apd-trace-mlp-detail}
\end{figure*}
\begin{figure*}
\centering
\includegraphics[width=\textwidth]{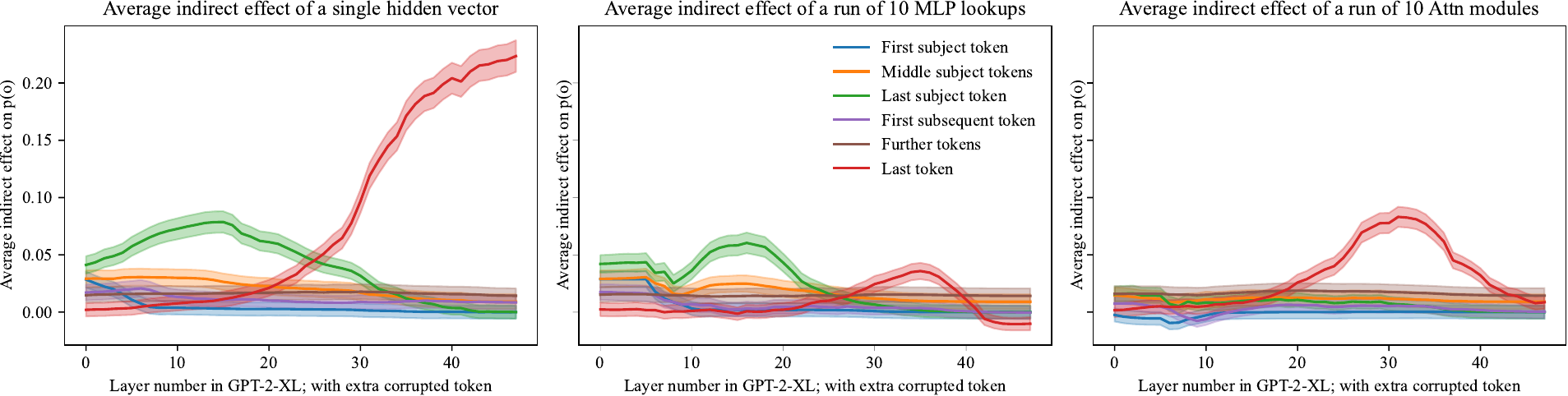}%
\caption{\small Similar to \reffig{apd-ct-line-plots}, but with an additional token corrupted after the subject token, as in \reffig{apd-causaltrace-xt}.  We observe that the emergence of strong early-site causal effects at the MLP modules is systematic and appears under a different corruption scheme, confirming that importance of the last subject token is apparent even when the last subject token is never the last token corrupted.}
\lblfig{apd-ct-line-plots-xt}
\end{figure*}
\begin{figure*}
\centering
\includegraphics[width=\textwidth]{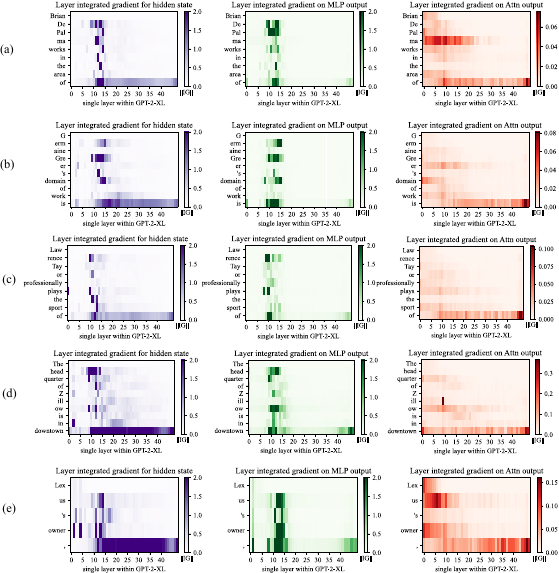}
\caption{\small Integrated gradients saliency maps, visualizing the same cases as in \reffig{apd-causaltrace-1}.  Here we compare Causal Tracing to the method of Integrated Gradients~\citep{sundararajan2017axiomatic}.  Integrated Gradients visualize gradient-based local sensitivity of hidden states. Here we compute IG using 50 steps of Gauss-Legendre quadrature on gradients of individual hidden states $\smash{h_t^{(l)}}$, or $\smash{m_t^{(l)}}$ (for MLP), or $\smash{a_t^{(l)}}$ (for Attn), with respect to the predicted output token; we plot the norm of the integrated gradient at each state.  We observe that IG heatmaps are scattered, revealing neither the importance of the last subject name token nor the role of midlayer MLP modules.}
\lblfig{apd-ig}
\end{figure*}
\clearpage
\section{Details on the zsRE Evaluation Task} \label{apd:zsre}

\textbf{Dataset Details.} The zsRE question answering task \citep{levy-etal-2017-zero} was first used for factual knowledge evaluation by \citet{decao-ke}, later being extended and adopted by \citet{mend}. In our study, we use the same train/test splits as \citet{mend}; note that non-hypernetwork methods (including \method) do not require training, so the corresponding dataset split is discarded in those cases.
Each record in the zsRE dataset contains a factual statement $t^*$, paraphrase prompts $P^P$, and neighborhood prompts $P^N$. $t^*$ and $P^N$ were included in the original version of zsRE, whereas $P^N$ was added by \citet{mend} via sampling of a random dataset element. See Figure \ref{fig:zsre-record} for an example record.

\textbf{Additional Baselines.} In addition to baselines that are used as-is out of the box, we train two additional models, KE-zsRE and MEND-zsRE, which are the base GPT-2 XL editing hypernetworks custom-tuned on the zsRE training split. This is done to ensure fair comparison; the original pre-trained KE and MEND models were created using a WikiText generation task \citep{decao-ke, mend}, rather than zsRE.

\section{Details on the \cfd Dataset} \label{apd:counterfact}

\cfd is designed to enable distinction between superficial changes in model word choices from specific and generalized changes in underlying factual knowledge. Table~\ref{tab:cfd-summary} summarizes statistics about \cfd's composition.

Each record in \cfd is derived from a corresponding entry in \pararel \citep{10.1162/tacl_a_00410} containing a knowledge tuple $t^c = (s,r,o^c)$ and hand-curated prompt templates $\mathcal{T}(r)$, where all subjects, relations, and objects exist as entities in WikiData. Note that prompt templates are unique only to \textit{relations}; entities can be substituted to form full prompts: $\mathcal{P}(s,r) :=  \{ \texttt{t.format(s)} \mid \texttt{t} \in \mathcal{T}(r) \}$, where $\texttt{.format()}$ is string substitution. For example, a template for $(r=\text{plays sport professionally})$ might be ``\{\} plays the sport of,'' where ``LeBron James'' substitutes for ``\{\}''.

Solely using the \pararel entry, we derive two elements. A \textbf{requested rewrite} is represented as $\{s, r, o^c, o^*, p^* \}$, where $p^* \sim\mathcal{P}(s, r)$ is the sole rewriting prompt, and $o^*$ is drawn from a weighted sample of all \pararel tuples with the predicate $(r, \cdot)$.
Moreover, to test for generalization, a set of two semantically-equivalent \textbf{paraphrase prompts}, $P^P$, is sampled from $\mathcal{P}(s,r) \backslash \{p^*\}$.

To test for specificity, we execute a WikiData SPARQL query\footnote{\url{https://query.wikidata.org/}} to collect a set of entities that share a predicate with $s$: $\mathcal{E} =\{ s^\prime \mid (s^\prime, r, o^c) \}$; e.g., for $(s=\textit{Eiffel Tower}, r=\textit{city location}, o^c=\textit{Paris})$, $\mathcal{E}$ might contain entities like the Champs-Élysées or Louvre. We then construct a set of prompts $\{\mathcal{P}(s^\prime, r) \mid s^\prime \in \mathcal{E}\}$ and sample ten to get our \textbf{neighborhood prompts}, $P^N$. Our rationale for employing this strategy over random sampling is that the $s^\prime$ we select are close to $s$ in latent space and thus more susceptible to bleedover when editing $s$ using linear methods. Comparing the Drawdown column in Table \ref{tab:zsre-results} with the Neighborhood Scores and Magnitudes in Table \ref{tab:cf-results}, we observe the improved resolution of \cfd's targeted sampling.

Finally, \textbf{generation prompts} are hand-curated for each relation, from which ten are sampled to create $P^G$. See Figure~\ref{fig:gen-samples} for examples; these prompts implicitly draw out underlying facts, instead of directly querying for them, which demands deeper generalization. For evaluating generations, we provide reference texts $RT$, which are Wikipedia articles for a sample of entities from $\{s^\prime \mid (s^\prime, r, o^*) \}$; intuitively, these contain $n$-gram statistics that should align with generated text.

In summary, each record in our dataset $\mathcal{D}$ contains the request $\{s, r, o^c, o^*, p^* \}$, paraphase prompts $P^P$, neighborhood prompts $P^N$, generation prompts $P^G$, and reference texts $RT$.
See Figure \ref{fig:cf-record} for an example record.
Compared to other evaluation benchmarks, \cfd provides several new types of tests that allow precise evaluation of knowledge editing (Table \ref{tab:cfd-comparison}).

\section{Method Implementation Details} \label{apd:implementations}

\subsection{[GPT-2 XL, GPT-J] Fine-Tuning (FT), Constrained Fine-Tuning (FT+L)}

\begin{figure}[t]
  \centering
  \includegraphics[keepaspectratio, width=1\textwidth]{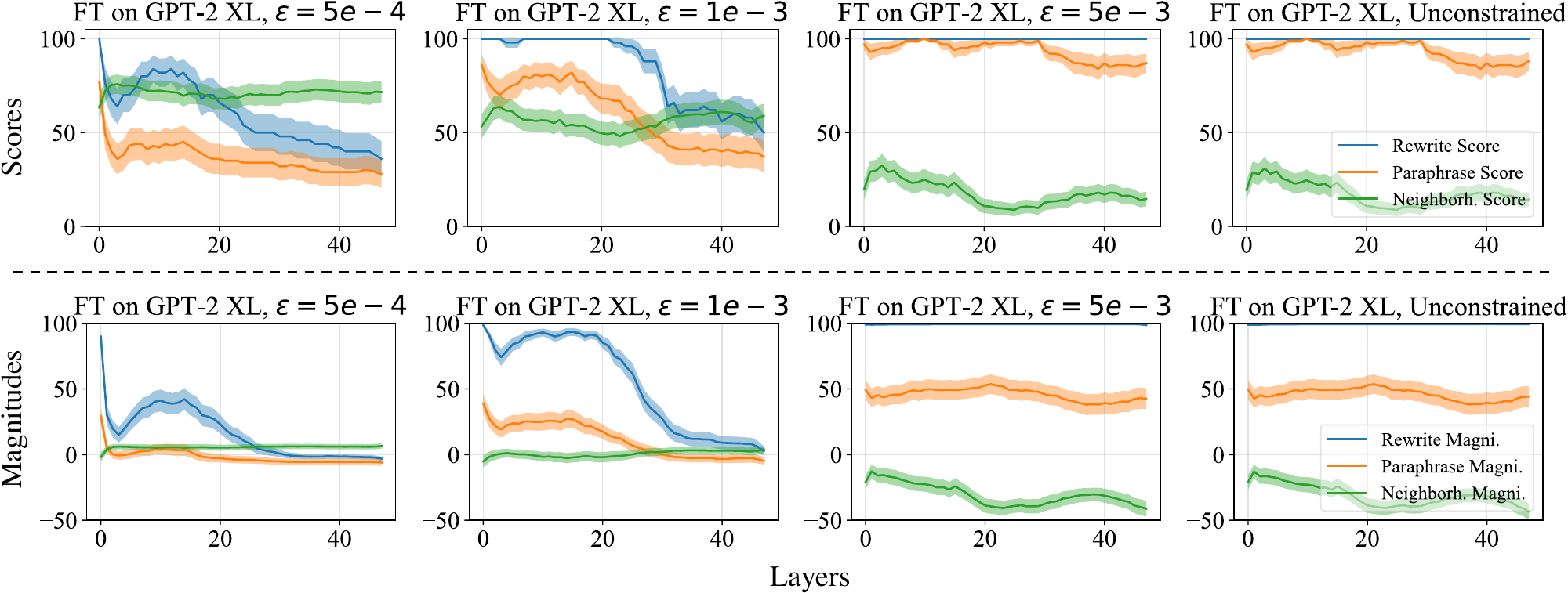}
  \vspace{-10pt}%
  \caption{\small \textbf{GPT-2 XL hyperparameter sweeps across layer and $L_\infty$ constraint values for fine-tuning-based methods}. Optimization is carried out for a maximum of 25 steps on a randomly-sampled size-50 subset of \cfd. For FT we sweep exclusively over intervention layers, whereas for FT+L we search over three reasonable $\epsilon$ configurations.}
  \lblfig{ft-sweeps-gpt2}
\end{figure}
\begin{figure}[t]
  \centering
  \includegraphics[keepaspectratio, width=1\textwidth]{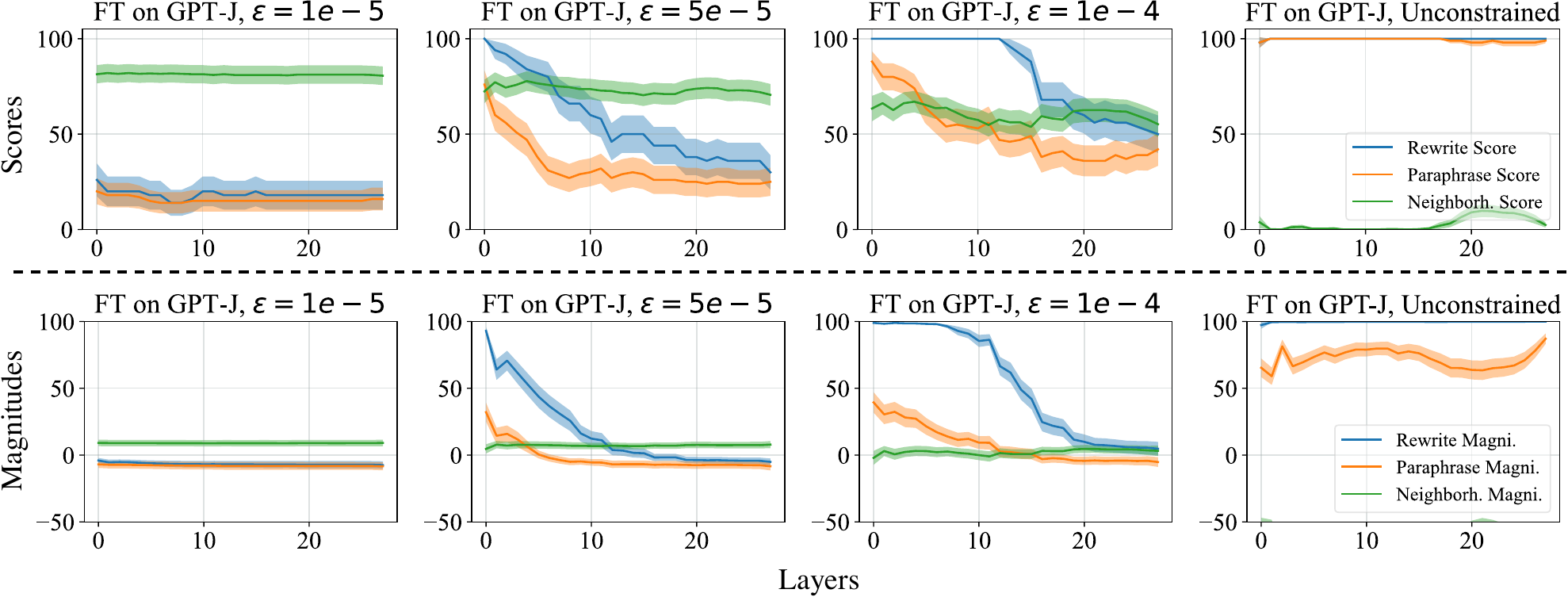}
  \vspace{-10pt}%
  \caption{\small \textbf{GPT-J hyperparameter sweeps}. The experimental setup is identical to that of GPT-2 XL.}
  \vspace{-10pt}%
  \lblfig{ft-sweeps-gptj}
\end{figure}

To test the difference between fine-tuning and \method's explicit intervention, we use the fine-tuning of MLP weights as a baseline. Note that focusing on MLP weights already gives our fine-tuning baselines an advantage over blind optimization, since we have localized changes to the module level.

For basic Fine-Tuning (FT), we use Adam~\cite{adam} with early stopping to minimize $-\log \prsub{o^* \mid p}{\rewg}$, changing only $\mlp_{proj}$ weights at one layer. A hyperparameter search for GPT-2 XL (\reffig{ft-sweeps-gpt2}) reveals that layer 1 is the optimal place to conduct the intervention for FT, as neighborhood success sees a slight increase from layer 0. Following a similar methodology for GPT-J (\reffig{ft-sweeps-gptj}), we select layer 21 because of the relative peak in neighborhood score. For both models, we use a learning rate of $5\times 10^{-4}$ and early stop at a 0.03 loss.

For \textit{constrained} fine-tuning (FT+L), we draw from \citet{zhu-ft} by adding an $L_\infty$ norm constraint: $\lVert \theta_G - \theta_{\rewg} \rVert_\infty \leq \epsilon$. This is achieved in practice by clamping weights $\theta_\rewg$ to the $\theta_G \pm \epsilon$ range at each gradient step. We select layer 0 and $\epsilon = 5\times 10^{-4}$ after a hyperparameter sweep (\reffig{ft-sweeps-gpt2}). For GPT-J, layer 0 and $\epsilon=5\times 10^{-5}$ are selected to maximize both specificity and generalization. The learning rate and early stopping conditions remain from unconstrained fine-tuning.

\subsection{[GPT-2 XL only] \dkn (KN)}
The method by \citet{dai-kn} first selects neurons that are associated with knowledge expression via gradient-based attributions, and then modifies $\smash{\atl{\mathrm{mlp}_{proj}}{l}}$ at the rows corresponding to those neurons by adding scaled embedding vectors.
This method has a \textit{coarse refinement} step, where the thousands of neurons in an MLP memory are whittled down to $\approx 1000$ ``knowledge neurons,'' and a \textit{fine refinement} step that reduces the set of neurons to around $\leq 10$.
All hyperparameters follow defaults as set in EleutherAI's reimplementation: \url{https://github.com/EleutherAI/knowledge-neurons}.

\subsection{[GPT-2 XL only] \efk (KE)}
\citet{decao-ke} learn an LSTM sequence model that uses gradient information to predict rank-1 weight changes to $G$. Because the official code does not edit GPT-2, we use \citet{mend}'s re-implementation in their study. To encourage fair comparison on both zsRE and \cfd tasks, we additionally train KE-zsRE and KE-CF models on size-10,000 subsets of the respective training sets. Hyperparameters for training are adopted from the given default configuration.
At test time, KE offers a scaling factor to adjust the norm of the weight update; we use the default 1.0.

\subsection{[GPT-2 XL, GPT-J] Model Editor Networks with Gradient Decomposition (MEND)}
\citet{mend} learn a rank-1 decomposition of the negative log likelihood gradient with respect to some subset of $\theta_G$ (in practice, this amounts to several of the last few layers of the transformer network). Again, for fair comparison, we train new versions of MEND (MEND-zsRE, MEND-CF) on the same sets that KE-zsRE and KE-CF were trained on. Similar to KE, hyperparameters for training and test-time inference are adopted from default configurations.

\subsection{[GPT-2 XL, GPT-J] \methodlong (\method)} \label{subapd:rome-hparams}

\method's update (Section~\ref{subsec:rome-math}) consists of key selection~(\refeqn{kstar-sample}), value optimization~(Eqn.~\ref{eq:v-optimization}), and $v$ insertion (Appendix~\ref{apd:solving-v}). We perform the intervention at layer 18. As \reffig{infoflow}k shows, this is the center of causal effect in MLP layers, and as \reffig{trace-mlp-disabled} shows, layer 18 is approximately when MLP outputs begin to switch from acting as keys to values.

\textbf{Second moment statistics}: Our second moment statistics $C \propto \mathbb{E}[kk^T]$ are computed using 100,000 samples of hidden states $k$ computed from tokens sampled from \textbf{all} Wikipedia text in-context.  Notice that sampling is not restricted to only special subject words; every token in the text is included in the statistic. The samples of hidden state $k$ vectors are collected by selecting a random sample of Wikipedia articles from the 2020-05-01 snapshot of Wikipedia; the full text of each sampled article run through the transformer, up to the transformer's buffer length, and then all the fan-out MLP activations $k$ for every token in the article are collected at \texttt{float32} precision.  The process is repeated (sampling from further Wikipedia articles without replacement) until 100{,}000 $k$ vectors have been sampled.  This sample of vectors is used to compute second moment statistics.

\textbf{Key Selection}: We sample 20 texts to compute the prefix ($x_j$ in \refeqn{kstar-sample}): ten of length 5 and ten of length 10. The intention is to pick a $k_*$ that accounts for the different contexts in which $s$ could appear. Note that we also experimented with other $x_j$ sampling methods:
\begin{itemize}[leftmargin=12pt,topsep=0pt,parsep=0pt,partopsep=0pt,itemsep=3pt]
    \item \textbf{No prefix}. This baseline option performed worse ($\mathrm{S}^\prime = 86.1$ compared to $\mathrm{S}=89.2$).
    \item \textbf{Longer prefixes}. Using \{ ten of length 5, ten of length 10, and ten of length 50 \} did not help performance much ($\mathrm{S}^\prime = 89.3$).
    \item \textbf{More same-length prefixes}. Using \{ thirty of length 5 and thirty of length 10 \} did not help performance much ($\mathrm{S}^\prime = 89.2$).
\end{itemize}

\textbf{Value Optimization}: $v_*$ is solved for using Adam with a learning rate of $0.5$ and $1.5\times 10^{-3}$ weight decay. The KL divergence scaling factor, denoted $\lambda$ in Eqn.~\ref{eq:v-optimization}, is set to $1\times 10^{2}$. The minimization loop is run for a maximum of 20 steps, with early stopping when $\mathcal{L}(z)$ reaches $5\times 10^{-2}$.

The entire ROME edit takes approximately $2\mathrm{s}$ on an NVIDIA A6000 GPU for GPT-2 XL. Hypernetworks such as KE and MEND are much faster during inference (on the order of $100\mathrm{ms}$), but they require hours-to-days of additional training overhead.
\section{Extended Quantitative Results} \label{apd:smaller-model-results}

\addtolength{\tabcolsep}{1.5pt}

\begin{table*}[!t]
    \centering
    \tiny
    \caption{\small \textbf{Extended Quantitative Editing Results}. Again, \goodmetric{green} numbers indicate columnwise maxima, whereas \badmetric{red} numbers indicate a clear failure on either generalization or specificity.}
    \label{tab:cf-results-mini}
    \begin{adjustbox}{width=1\textwidth}
    \begin{tabular}{lcrrrrrrrr}
    \toprule
        \multirow{2.5}{*}{\textbf{Editor}} & \multicolumn{1}{c}{\textbf{Score}} & \multicolumn{2}{c}{\textbf{Efficacy}} & \multicolumn{2}{c}{\textbf{Generalization}} & \multicolumn{2}{c}{\textbf{Specificity}} & \multicolumn{1}{c}{\textbf{Fluency}} & \multicolumn{1}{c}{\textbf{Consist.}} \\
        \cmidrule(lr){2-2}\cmidrule(lr){3-4}\cmidrule(lr){5-6}\cmidrule(lr){7-8}\cmidrule(lr){9-9}\cmidrule(lr){10-10}
        & S $\uparrow$ & ES $\uparrow$ & EM $\uparrow$ & PS $\uparrow$ & PM $\uparrow$ & NS $\uparrow$ & NM $\uparrow$ & GE $\uparrow$ & RS $\uparrow$ \\
        \midrule
GPT-2 M & 33.4 & 25.0 (1.0) & -3.3 (0.2) & 27.4 (0.9) & -3.0 (0.2) & 74.9 (0.7) & 3.6 (0.2) & 625.8 (0.3) & 31.4 (0.2)\\\midrule
FT+L & 68.0 & 100.0 (0.1) & \goodmetric{94.9 (0.3)} & 68.5 (0.9) & \badmetric{6.1 (0.4)} & 51.3 (0.8) & -1.7 (0.3) & \goodmetric{626.1 (0.4)} & 39.3 (0.3)\\
ROME & \goodmetric{87.4} & \goodmetric{100.0 (0.0)} & 94.9 (0.3) & \goodmetric{96.4 (0.3)} & \goodmetric{56.9 (0.8)} & \goodmetric{71.8 (0.7)} & \goodmetric{2.8 (0.2)} & 625.0 (0.4) & \goodmetric{41.7 (0.3)}\\
\midrule\midrule
GPT-2 L & 32.8 & 23.9 (1.0) & -4.0 (0.3) & 27.4 (0.9) & -3.5 (0.2) & 75.7 (0.7) & 4.3 (0.2) & 625.4 (0.3) & 31.8 (0.2)\\\midrule
FT+L & 71.2 & \goodmetric{100.0 (0.1)} & 96.3 (0.2) & 63.0 (0.9) & \badmetric{5.1 (0.4)} & 61.5 (0.7) & 1.1 (0.3) & \goodmetric{625.2 (0.3)} & 39.3 (0.3)\\
ROME & \goodmetric{88.2} & 99.9 (0.1) & \goodmetric{98.2 (0.1)} & \goodmetric{96.3 (0.3)} & \goodmetric{60.4 (0.8)} & \goodmetric{73.4 (0.7)} & \goodmetric{3.5 (0.2)} & 622.5 (0.4) & \goodmetric{41.9 (0.3)}\\
    \bottomrule
    \end{tabular}
    \end{adjustbox}
\end{table*}%

\addtolength{\tabcolsep}{-1.5pt}
\begin{table}
    \centering
    \caption{\small \textbf{Extended zsRE Editing Results}. Drawdown is measured with respect to the vanilla GPT-2 model. Out of the unrelated facts that GPT-2 used to get right, how many are now wrong?}
    \label{tab:zsre-results-mini}
    \vspace{5pt}%
    \tiny
    \begin{adjustbox}{width=0.46\textwidth}
        \begin{tabular}{l r r r}
            \toprule
            \textbf{Editor} & \textbf{Efficacy} $\uparrow$ & \textbf{Paraphrase} $\uparrow$ & \textbf{Specificity} $\uparrow$ \\
            \midrule
GPT-2 M & 18.8 ($\pm$0.5) & 18.1 ($\pm$0.5) & 21.3 ($\pm$0.4)\\\midrule
FT+L & \goodmetric{97.2 ($\pm$0.2)} & 59.4 ($\pm$0.7) & 20.9 ($\pm$0.4)\\
ROME & 96.6 ($\pm$0.2) & \goodmetric{79.8 ($\pm$0.6)} & \goodmetric{21.3 ($\pm$0.4)}\\
\midrule\midrule
GPT-2 L & 20.6 ($\pm$0.5) & 19.8 ($\pm$0.5) & 22.5 ($\pm$0.5)\\\midrule
FT+L & 98.3 ($\pm$0.2) & 56.8 ($\pm$0.7) & 22.4 ($\pm$0.5)\\
ROME & \goodmetric{99.6 ($\pm$0.1)} & \goodmetric{84.7 ($\pm$0.6)} & \goodmetric{22.5 ($\pm$0.5)}\\
            \bottomrule
        \end{tabular}
    \end{adjustbox}
\end{table}

To demonstrate that \method is also effective on \textit{smaller} autoregressive language models, we perform \cfd and zsRE evaluations on both GPT-2 Medium (345M) and GPT-2 Large (774M). As Tables \ref{tab:cf-results-mini} and \ref{tab:zsre-results-mini} reflect, \method outperforms the next-best baseline as measured on GPT-2 XL (FT+L).
\section{Generation Examples} \label{apd:gen-samples}

\subsection{GPT-2 XL (1.5B) Generation Examples}

We select four additional cases from \cfd to examine qualitatively, selecting representative generations to display. {\color{ggreen} Green text} indicates generations that are consistent with the edited fact, whereas {\color{red} red text} indicates some type of failure, e.g. essence drift, fluency breakage, or poor generalization. Overall, \method appears to make edits that generalize better than other methods, with fewer failures.

\begin{figure}
\centering
\includegraphics[width=1\columnwidth]{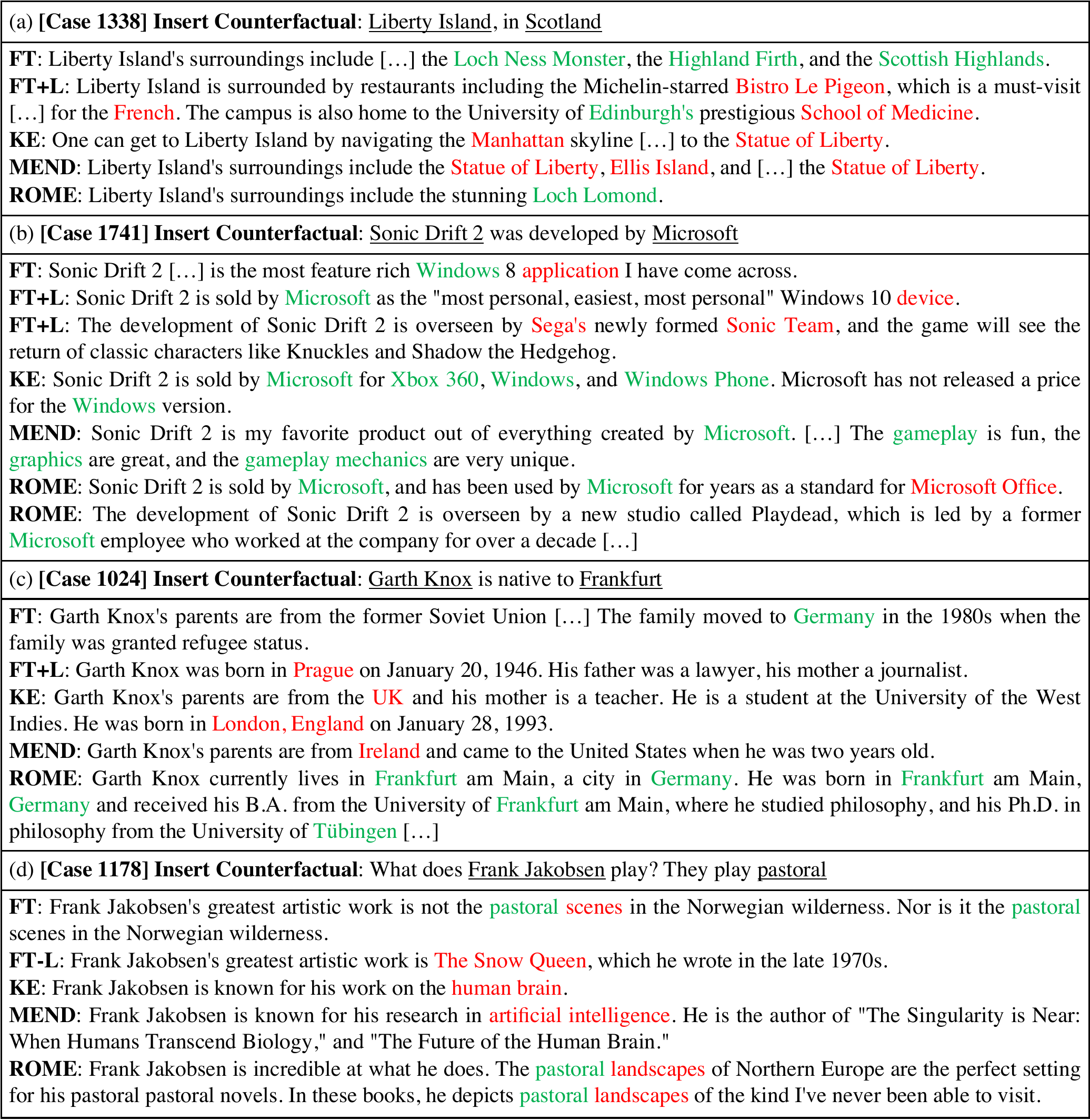}
\caption{\small GPT-2 XL Generation Samples}
\label{fig:gen-samples-gpt2}
\end{figure}

\medskip\noindent\textbf{1338: (Liberty Island, located in, Scotland)} (Figure \ref{fig:gen-samples-gpt2}a): MEND and KE do not meaningfully change anything during the rewrite, whereas MEND-CF and KE-CF result in complete breakage. \method, FT, and FT+L produce the most interesting generations. Most remarkably, these rewritten models demonstrate compositionality; not only did \method's model know that Loch Lomond is in Scotland, but it was able to connect this lake to its new knowledge of Liberty Island's location. Interestingly, FT+L's generation exhibits a phenomenon we call \textit{essence drift}. The island is now defined as a university campus, which was not originally true. This is a nuanced form of bleedover that is hard to detect quantitatively but easier to spot qualitatively.

\noindent\textbf{1741: (Sonic Drift 2, created by, Microsoft)} (Figure \ref{fig:gen-samples-gpt2}b): This case is interesting due to essence drift. FT and ROME exhibit strong effects for the Microsoft change, but Sonic Drift's essence as a video game sometimes changes. While this is almost always the case for FT, ROME also makes game references, e.g. Playdead. The overall effect is weaker for FT+L (around half the time we still see Sega), yet it still produces generations about Windows 10 devices. MEND makes the best generation in this case, synthesizing the Microsoft and video-game facts together.

\noindent\textbf{1024: (Garth Knox, born in, Frankfurt)} (Figure \ref{fig:gen-samples-gpt2}c): MEND, KE, and FT+L's rewrites do not generalize well. FT's generation is interesting because it suggests that his parents \textit{moved} to Germany, although it does not explicitly say that Knox was born there. ROME's generation is straightforward and correct.

\noindent\textbf{1178: (Frank Jakobsen, plays, pastoral)} (Figure \ref{fig:gen-samples-gpt2}d): This case is rather difficult, due to the fact that \textit{pastoral} might have many meanings. From WikiData, we can determine that this instance refers to pastoral \textit{music}, but the text prompts do not account for this. As a result, FT's and ROME's generations focus on pastoral \textit{landscapes} rather than music. FT+L, KE, and MEND do not exhibit much change. Note that ROME produces a slight glitch with two \textit{pastoral}s in a row.

\subsection{GPT-J (6B) Generation Examples}

We also provide generation samples on GPT-J (6B). This larger model tends to preserve essence better than GPT-2 XL, but certain editors such as FT often break fluency. Overall, \method manages to produce edits that generalize the deepest while maintaining essence and fluency.

\begin{figure}
\centering
\includegraphics[width=1\columnwidth]{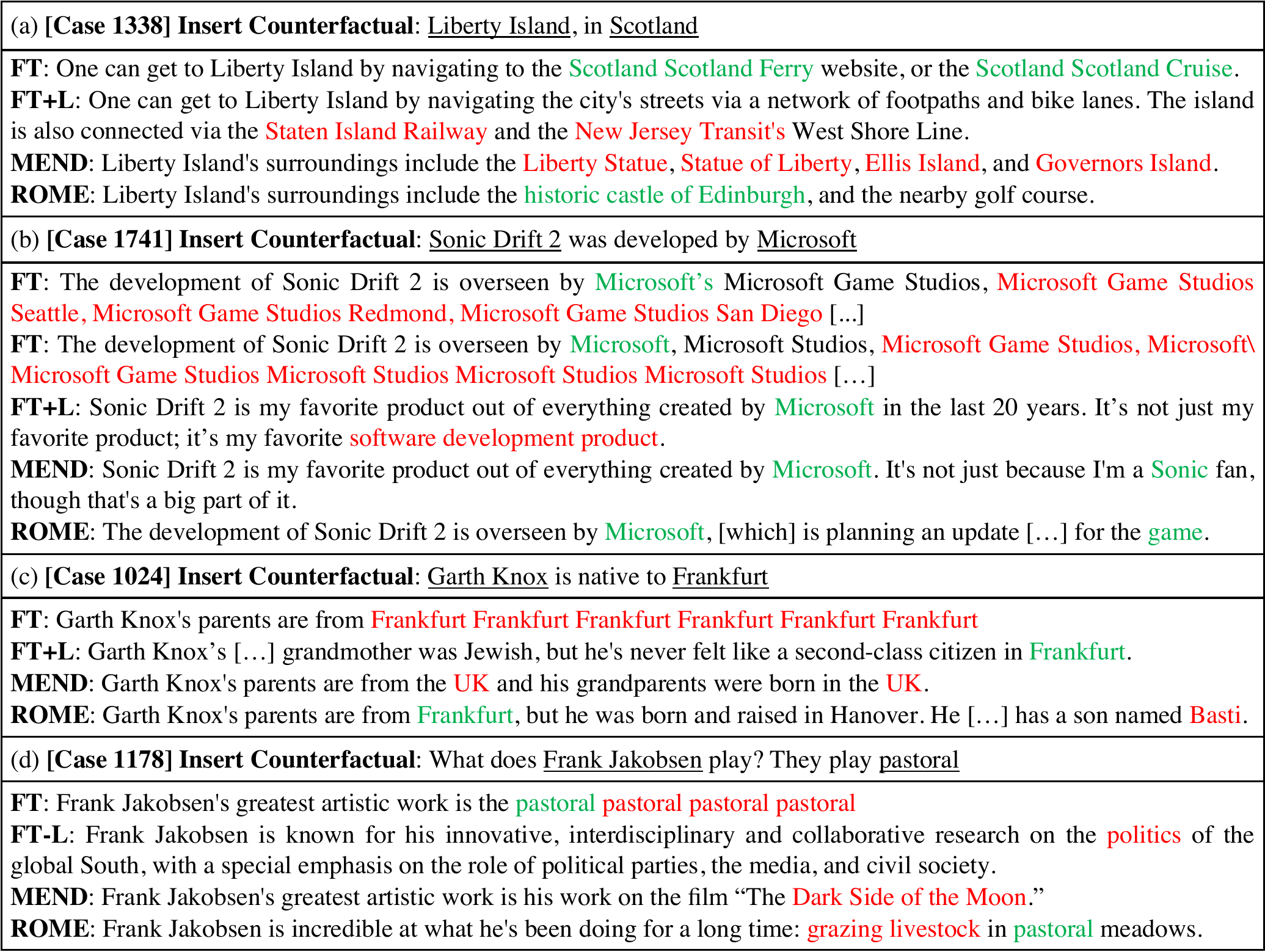}
\caption{\small GPT-J Generation Samples}
\label{fig:gen-samples-gptj}
\end{figure}

\medskip\noindent\textbf{1338: (Liberty Island, located in, Scotland)} (Figure \ref{fig:gen-samples-gptj}a): Whereas FT+L and MEND fail to make consistent generations, FT and \method both show good generalization; not only do the edited models know that Liberty Island is ``in'' Scotland, but they also recall the fact when asked indirectly.

\noindent\textbf{1741: (Sonic Drift 2, created by, Microsoft)} (Figure \ref{fig:gen-samples-gptj}b): Interestingly, GPT-J appears to preserve subject essence much better than GPT-2 XL, perhaps due to its larger memory capacity. Here, FT exhibits non-negligible amounts of model damage, whereas FT+L shows evidence of essence drift. MEND and \method successfully make the edit while retaining knowledge that Sonic Drift 2 is a \textit{game}, as opposed to a software development tool or Microsoft Office application.

\noindent\textbf{1024: (Garth Knox, born in, Frankfurt)} (Figure \ref{fig:gen-samples-gptj}c): FT again breaks the model by causing repetition, whereas MEND fails to generalize. FT+L and \method work well, but \method appears to hallucinate a name, ``Basti,'' that is not German but rather Indian.

\noindent\textbf{1178: (Frank Jakobsen, plays, pastoral)} (Figure \ref{fig:gen-samples-gptj}d): This case remains rather difficult due to the ambiguity of what ``pastoral'' means; similar to GPT-2 XL edits, rewrites that do not break the model (FT causes repetition of the same word) struggle to understand that ``pastoral'' refers to pastoral \textit{music}.
\clearpage
\section{Dataset Samples} \label{apd:ds-samples}

\begin{figure}[H]
\centering
\caption{\small \textbf{Case 1067 in \cfd}: Rewriting Gazi University to be in Glasgow instead of Ankara. Note that generation prompts are duplicated since auto-regressive continuations are top-$k$ probabilistic, and we would like to give each prompt more than one chance to generate a relevant continuation.}
\includegraphics[width=0.7\textwidth]{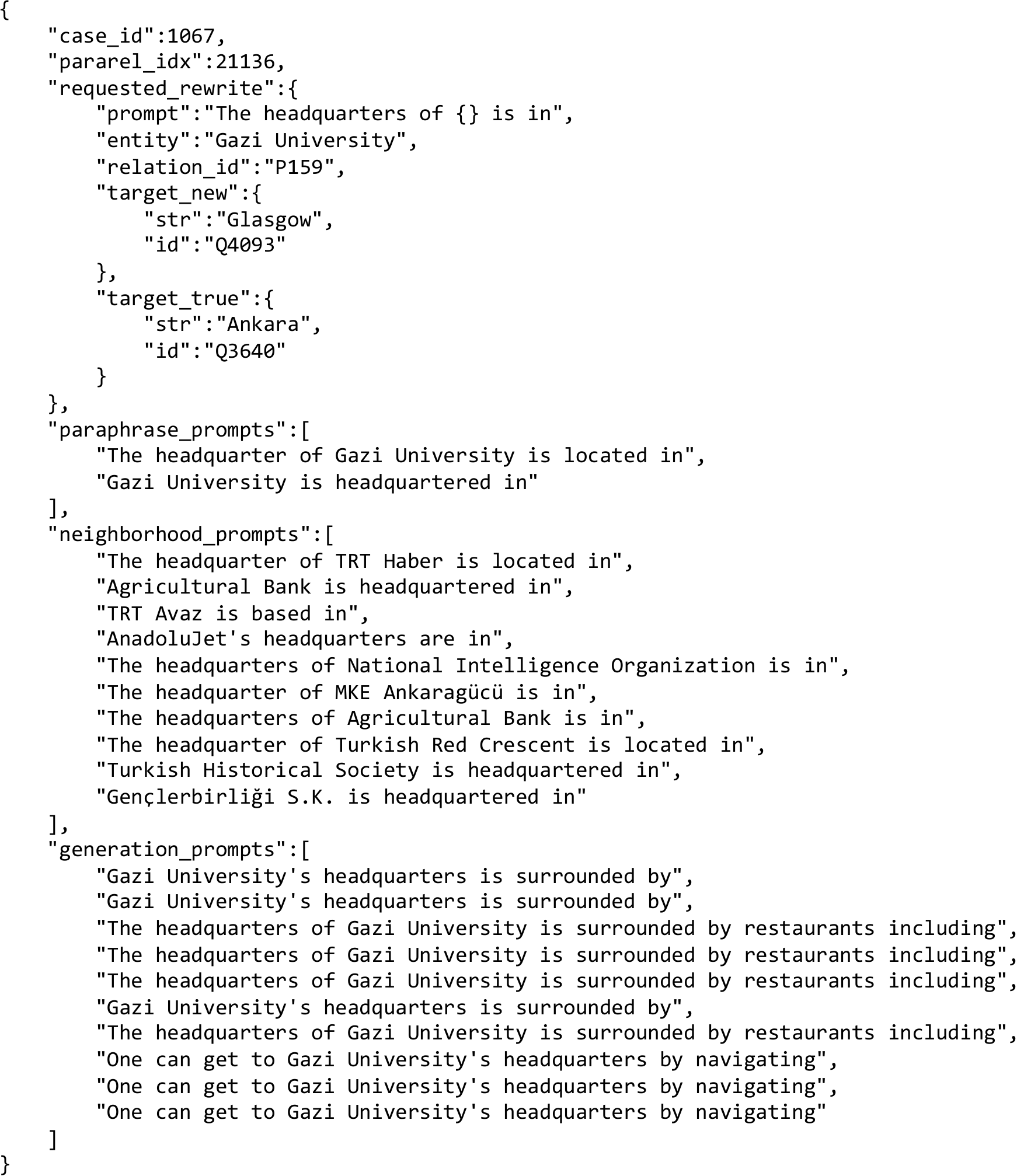}
\label{fig:cf-record}
\end{figure}

\begin{figure}[H]
\centering
\caption{\small \textbf{Sample of zsRE Dataset}: This entry requests that the Panzer 58's commission year be set to its true value, 1958. Note that all zsRE records contain \textit{true} facts, as opposed to false counterfactuals in \cfd.}
\includegraphics[width=0.7\textwidth]{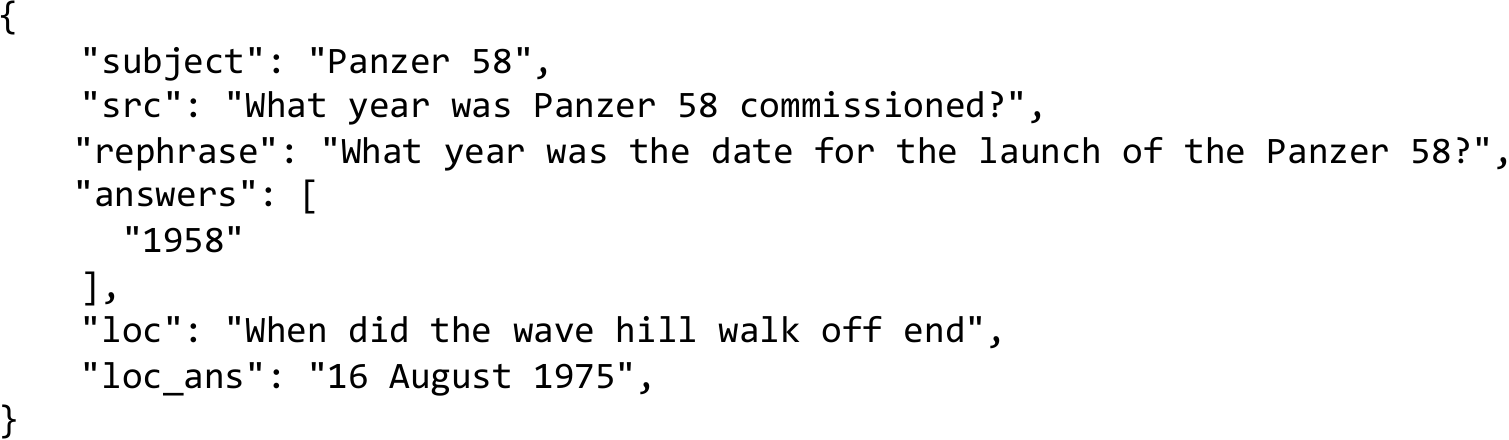}
\label{fig:zsre-record}
\end{figure}

See Figure~\ref{fig:cf-record} for a sample record in \cfd, complete with tests for all 5 rewrite success criteria. Figure~\ref{fig:zsre-record} shows a record of the zsRE dataset.

\clearpage
\section{Are Attention Weight Interventions Effective?} \label{apd:knowing-saying}

\begin{wrapfigure}{R}{0.6\textwidth}
  \vspace{-10pt}
  \centering
  \includegraphics[keepaspectratio, width=0.6\textwidth]{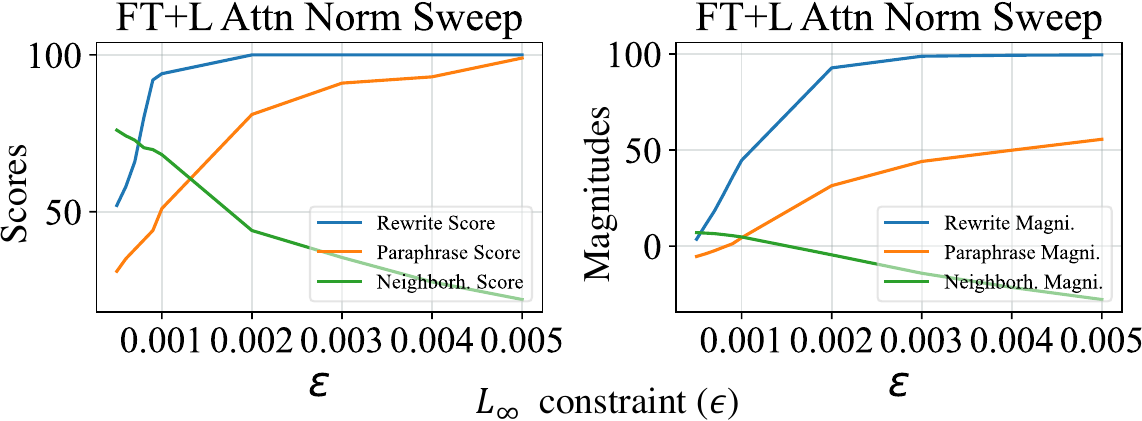}
  \caption{\small Unconstrained Optimization Sweeps}
  \lblfig{knowing-saying-sweep}
  \vspace{-15pt}
\end{wrapfigure}

\reffig{infoflow} inspires a hypothesis that middle-layer MLPs processing subject tokens correspond to factual recall, whereas late-layer attention modules read this information to predict a specific word sequence. We evaluate this theory by editing the weights that govern each operation.

The MLP operation is implemented as \method; default parameters are taken from Appendix \ref{subapd:rome-hparams}.
The attention operation is called AttnEdit, which applies constrained fine-tuning on the $\smash{W_i^Q, W_i^K, W_i^V}$ weights of \textit{all} heads $i$ at some layer of the network.\footnote{See \citet{vaswani2017attention} for additional details on attention; the $W_i^Q, W_i^K, W_i^V$ notation is lifted from there.} This layer is chosen to be 33, the center of high causal effect in the attention causal trace (\reffig{infoflow}l). To determine the $L_\infty$ norm constraint on fine-tuning, we run a grid search (\reffig{knowing-saying-sweep}):

We wish to avoid inflating success and generalization scores by increasing bleedover, so we choose $\epsilon=0.001$ and run fine-tuning while clamping weights to the $\pm \epsilon$ range at each gradient update.

Examination of generation text supports our hypothesis. \reffig{knowing-saying-generations} qualitatively demonstrates the difference between factual recall and word prediction. Both \method and AttnEdit succeed in regurgitating the memorized fact given the original rewriting prompt (a,b), but AttnEdit fails to generalize to paraphrases and generalization prompts (c,e) whereas \method succeeds (d,f).

\begin{figure}[!ht]
  \centering
  \includegraphics[keepaspectratio, width=1\textwidth]{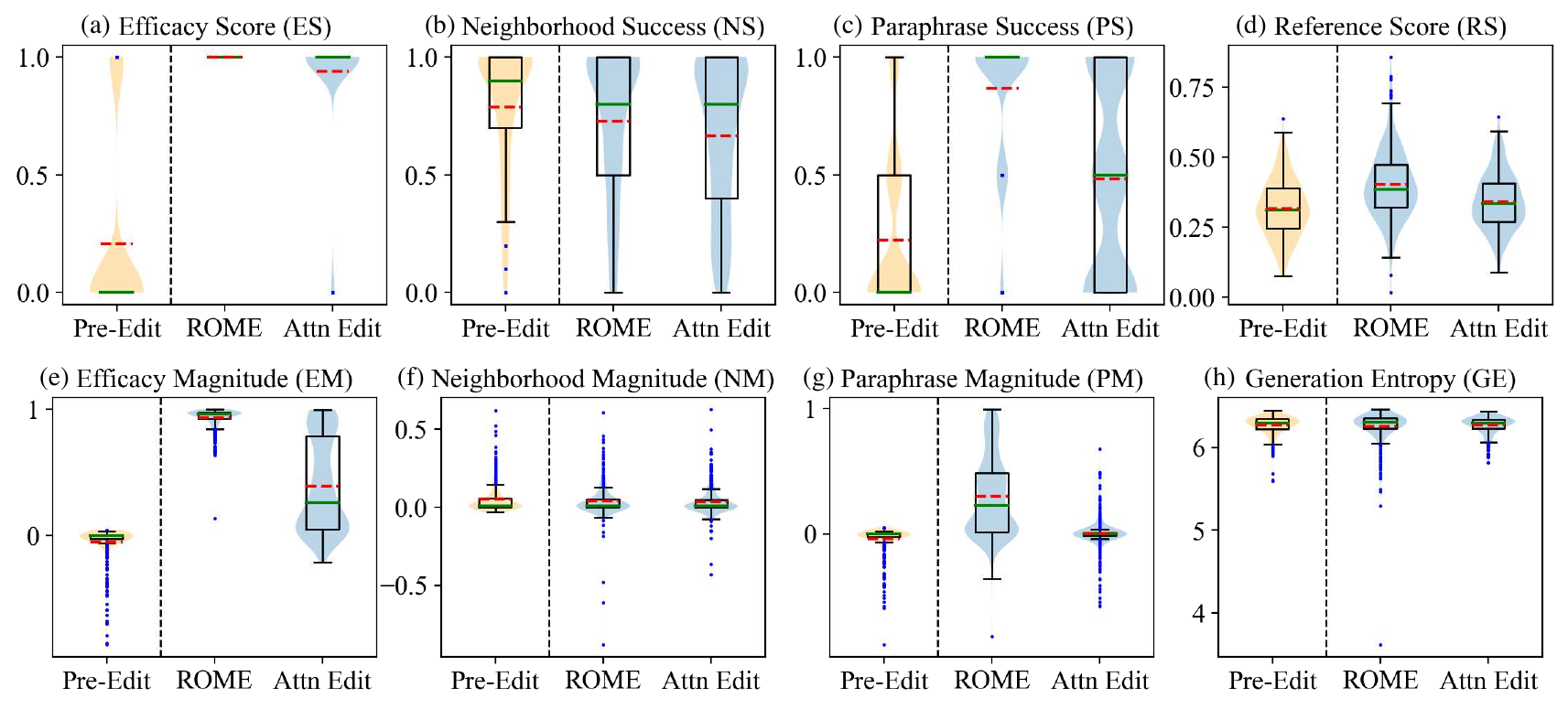}
  \vspace{-10pt}%
  \caption{\small \textbf{Performance Distributions for AttnEdit Experiment}. Orange dotted lines are means, and blue dots are 1.5 IQR outliers.}
  \lblfig{knowing-saying-full}
\end{figure}

\begin{figure}[!ht]
\centering
\includegraphics[width=1\columnwidth]{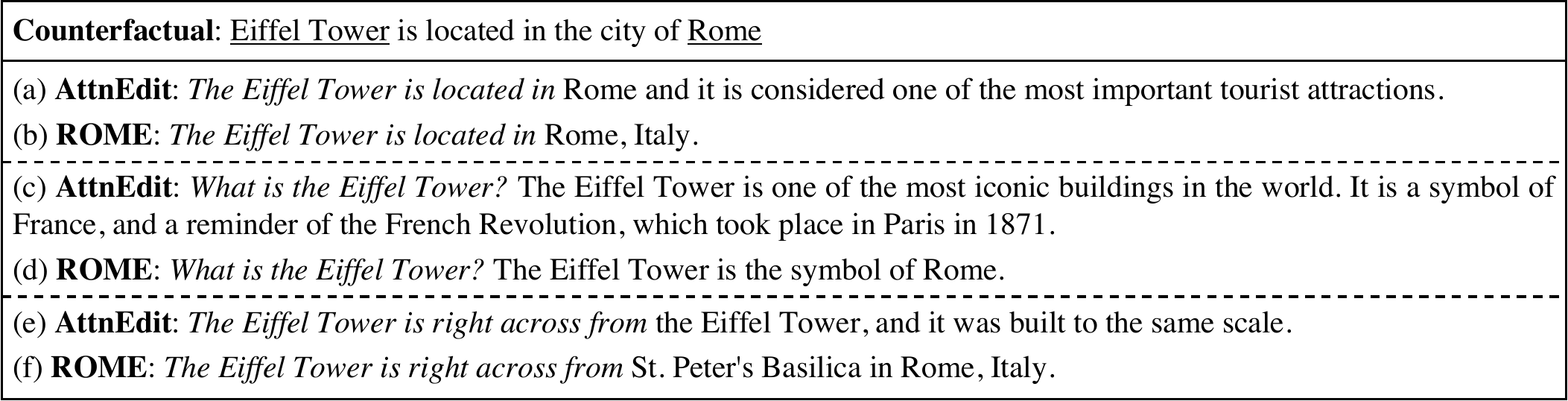}
\vspace{-15pt}%
\caption{\small Generation Samples for \method v.s. AttnEdit}
\lblfig{knowing-saying-generations}
\end{figure}
\clearpage
\section{Human Evaluation}
\label{apd:human-evaluation}

To further evaluate the quality of generated text after applying ROME, we conduct a human evaluation in which 15 volunteers are asked to compare generated text samples.  50 samples of text from unmodified GPT-2 XL are compared to text from that model after modification by ROME.  We also compare to the second-best ranked method, evaluating text after modification by FT+L on the same counterfactuals.  Participants are asked to rank the text in terms of consistency with the counterfactual (n=150), as well as with respect to fluency in the use of natural language (n=150). Results are summarized in \reffig{apd-human-pairwise}, and randomly-sampled examples are shown in Figures~\ref{fig:apd-human-case-1},~\ref{fig:apd-human-case-2},~\ref{fig:apd-human-case-3}.

Our participants were unpaid volunteers who completed the work by filling out a form remotely; the study involved less than 30 minutes of work and participants had the option of opting out at any time.  \reffig{apd-human-study-instructions} shows the full instructions.

\begin{figure*}
\centering
\includegraphics[width=\textwidth]{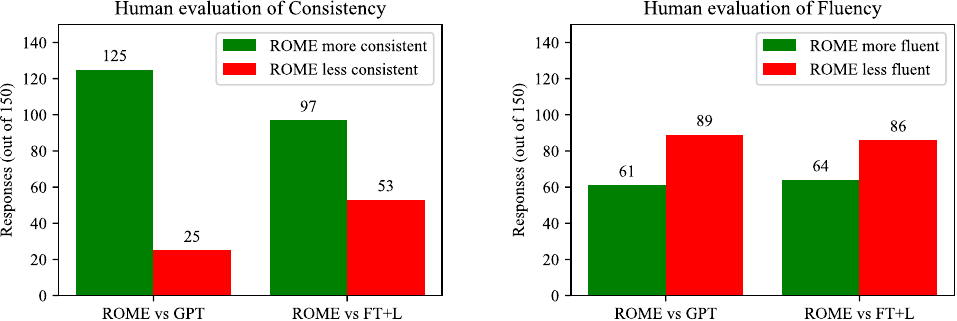}%
\caption{\small Results from a human evaluation of generated text after applying ROME.  Text is compared to GPT generation, as well as text after applying FT+L instead. Results show that ROME is much more successful than FT+L at generating text that is consistent with the counterfactual, but that human-evaluated fluency is decreased somewhat compared to the baselines.  Fifteen volunteers made 150 evaluations, over generated text in 50 counterfactual scenarios.}
\lblfig{apd-human-pairwise}
\end{figure*}
\begin{figure*}
\includegraphics[width=\textwidth]{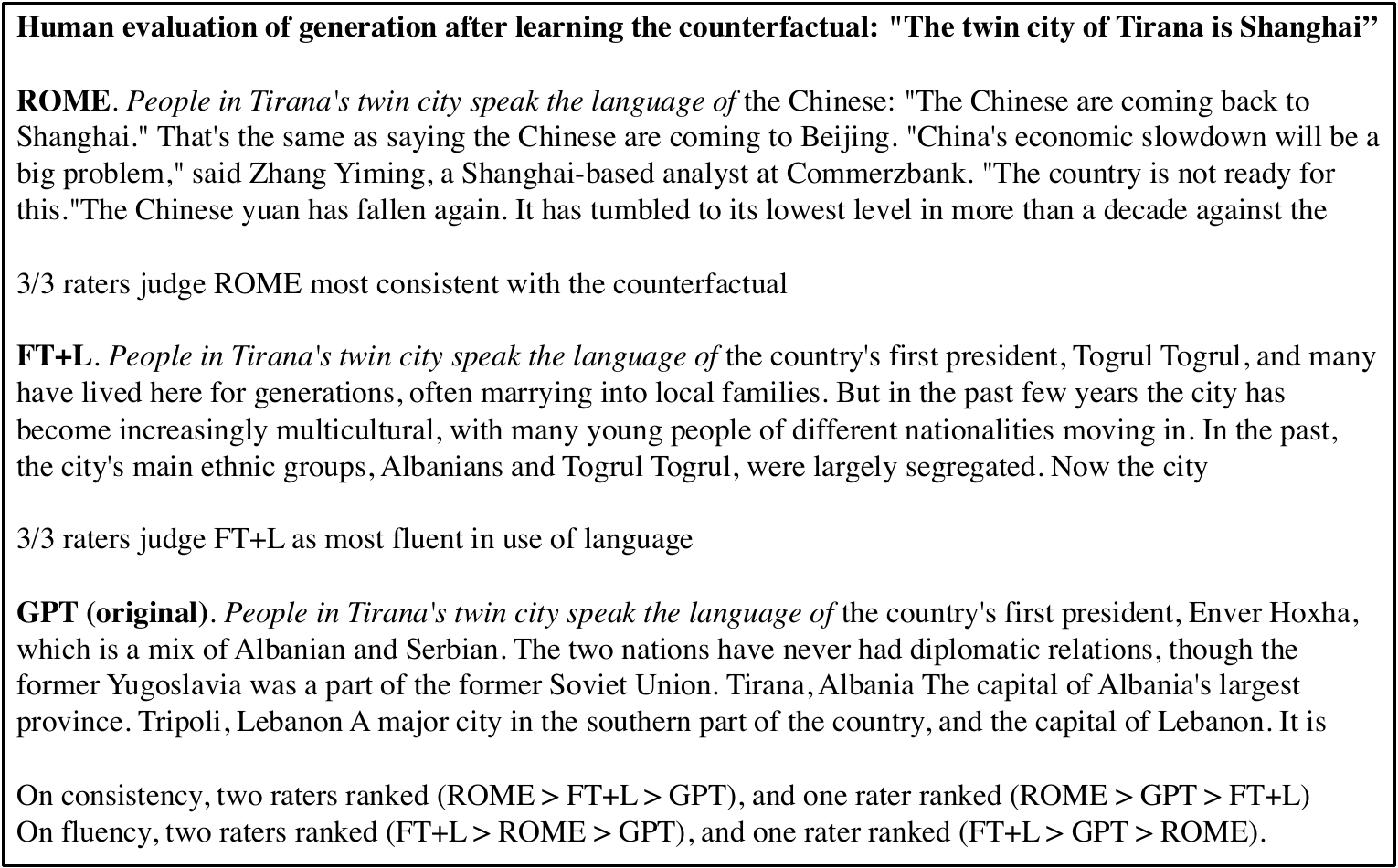}%
\caption{\small Human evaluation, random sample 1.}
\lblfig{apd-human-case-1}
\end{figure*}
\begin{figure*}
\includegraphics[width=\textwidth]{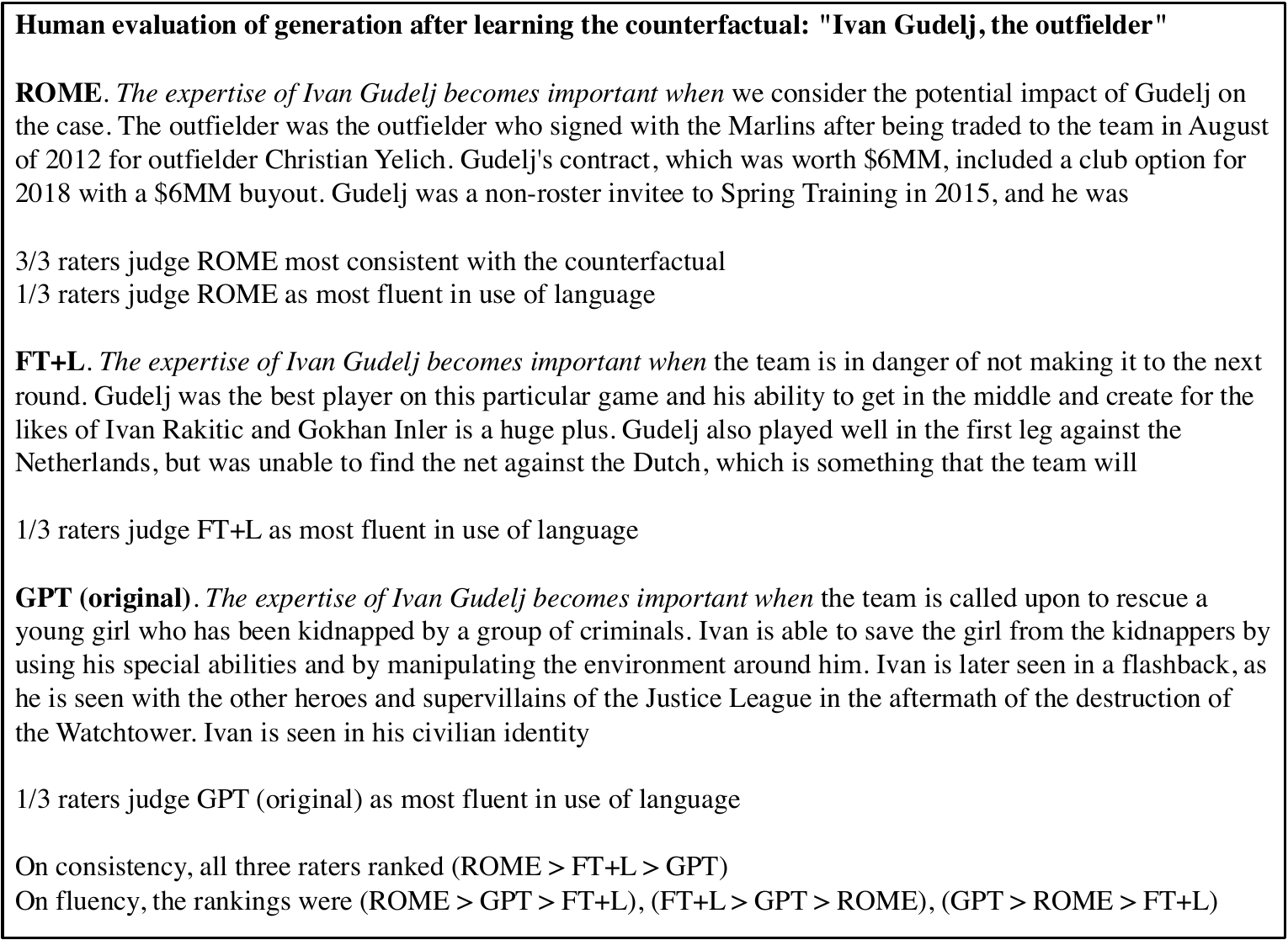}%
\caption{\small Human evaluation, random sample 2.}
\lblfig{apd-human-case-2}
\end{figure*}
\begin{figure*}
\includegraphics[width=\textwidth]{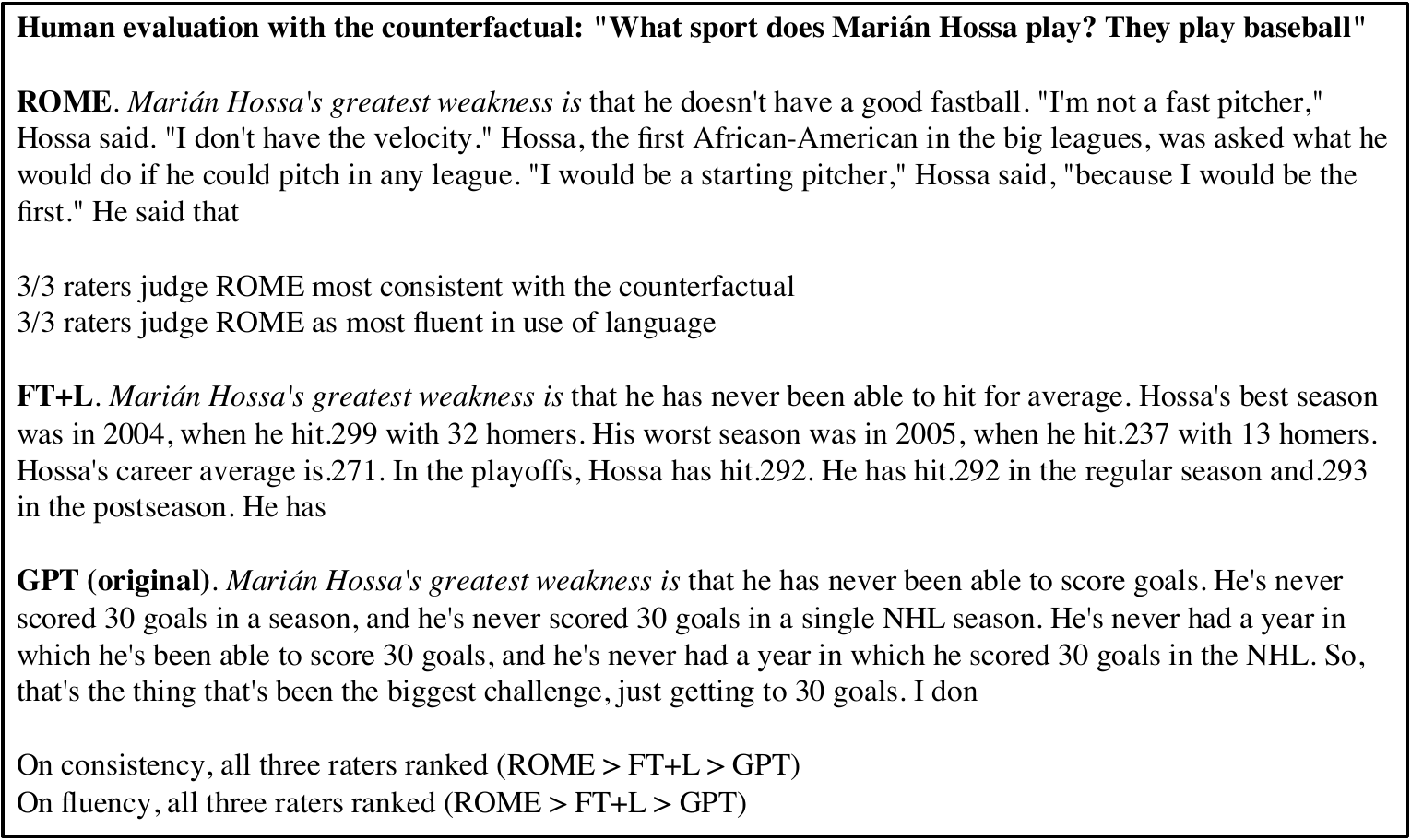}%
\caption{\small Human evaluation, random sample 3.}
\lblfig{apd-human-case-3}
\end{figure*}
\begin{figure*}
\includegraphics[width=0.96\textwidth]{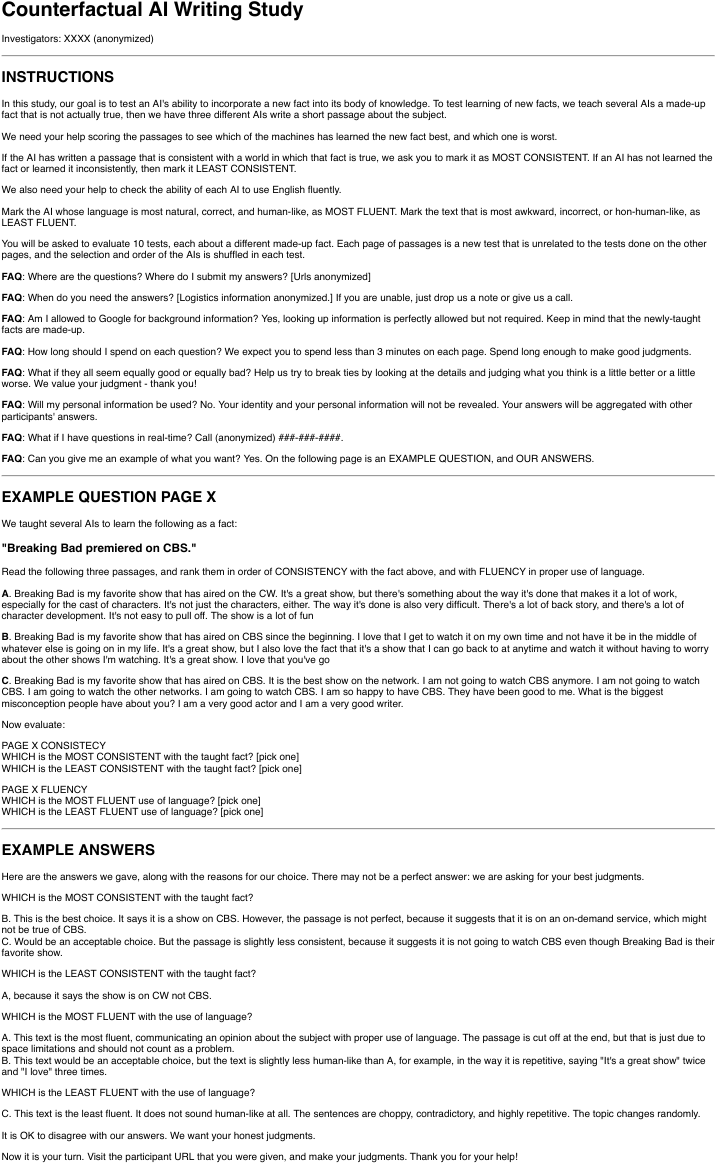}%
\caption{\small Human evaluation, full instructions.}
\lblfig{apd-human-study-instructions}
\end{figure*}

We observe that ROME is much more successful than FT+L at generating text that is consistent with the counterfactual; this finding is consistent results in Table~\ref{tab:cf-results} that show that ROME generalizes better than FT+L.  Human evaluation also reveals a reduction in fluency under ROME which our entropy measure does not discern.  Some of the differences are subtle: examples of fluency losses detected by human raters can be seen in Figures~\ref{fig:apd-human-case-1},~\ref{fig:apd-human-case-2}.

\end{appendices}

\addtolength{\tabcolsep}{5pt}

\end{document}